\documentclass{article}

\usepackage{float}  % for placing floating images better
\usepackage{dblfloatfix}
\usepackage{graphicx}
\usepackage{booktabs}
\usepackage{tikz}
\usetikzlibrary{arrows, arrows.meta, shapes, trees, positioning, calc, fit}
\usepackage[frozencache,cachedir=.]{minted}
\DeclareUnicodeCharacter{2293}{$\sqcup$}
\DeclareUnicodeCharacter{2264}{$\leq$}
\DeclareUnicodeCharacter{1EF3}{\`y}
\usepackage{hyphenat}
\usepackage{microtype}
\usepackage{graphicx}
\usepackage{subfigure}
\usepackage{booktabs} % for professional tables

\usepackage{hyperref}

\usepackage[accepted]{icml2024}

\usepackage{amsmath}
\usepackage{amssymb}
\usepackage{mathtools}
\usepackage{amsthm}

\usepackage[capitalize,noabbrev]{cleveref}

\theoremstyle{plain}

\theoremstyle{definition}

\theoremstyle{remark}

\usepackage[disable,textsize=tiny]{todonotes}

\begin{document}

\twocolumn[
\icmltitle{Graph2Tac: Online Representation Learning of Formal Math Concepts}



\icmlsetsymbol{equal}{*}
\icmlsetsymbol{workatihes}{†}
\icmlsetsymbol{workatihesibm}{‡}

\begin{icmlauthorlist}
\icmlauthor{Lasse Blaauwbroek}{equal,ihes}
\icmlauthor{Miroslav Olšák}{equal,cam,workatihes}
\icmlauthor{Jason Rute}{equal,ibm}
\icmlauthor{Fidel Ivan Schaposnik Massolo}{ihes}
\icmlauthor{Jelle Piepenbrock}{ru,ctu}
\icmlauthor{Vasily Pestun}{ihes,ibm,workatihesibm}
\end{icmlauthorlist}

\icmlaffiliation{ibm}{IBM Research}
\icmlaffiliation{ihes}{IHES}
\icmlaffiliation{cam}{University of Cambridge}
\icmlaffiliation{ru}{Radboud University}
\icmlaffiliation{ctu}{Czech Technical University in Prague
% Add these notes here so they appear at the end of the affiliation list
†Work performed at IHES
‡Work performed while at IHES and IBM Research until June 2022}

%\icmlaffiliation{yyy}{Department of XXX, University of YYY, Location, Country}
%\icmlaffiliation{comp}{Company Name, Location, Country}
%\icmlaffiliation{sch}{School of ZZZ, Institute of WWW, Location, Country}

\icmlcorrespondingauthor{Lasse Blaauwbroek}{lasse@blaauwbroek.eu}
\icmlcorrespondingauthor{Jason Rute}{jason.rute@ibm.com}

\icmlkeywords{Theorem Proving, Coq, Rocq, Representation Learning}

\vskip 0.3in
]



\printAffiliationsAndNotice{\icmlEqualContribution} % otherwise use the standard
% text.

\begin{abstract}
  In proof assistants, the physical proximity between two formal mathematical
  concepts is a strong predictor of their mutual relevance. Furthermore, lemmas
  with close proximity regularly exhibit similar proof structures. We show that
  this \textit{locality} property can be exploited through online learning
  techniques to obtain solving agents that far surpass offline learners when
  asked to prove theorems in an unseen mathematical setting. We extensively
  benchmark two such online solvers implemented in the Tactician platform for
  the Coq proof assistant: First, Tactician's online $k$-nearest neighbor
  solver, which can learn from recent proofs, shows a $1.72\times$ improvement in
  theorems proved over an offline equivalent. Second, we introduce a graph
  neural network, Graph2Tac, with a novel approach to build hierarchical
  representations for new definitions. Graph2Tac's online definition task
  realizes a $1.5\times$ improvement in theorems solved over an offline baseline. The
  $k$-NN and Graph2Tac solvers rely on orthogonal online data, making them
  highly complementary. Their combination improves $1.27\times$ over their
  individual performances. Both
  solvers outperform all other general-purpose provers for Coq, including
  CoqHammer, Proverbot9001, and a transformer baseline by at least
  $1.48\times$ and are available for practical use by end-users.
\end{abstract}

% \begin{abstract}
%   When deploying AI tools for interactive theorem proving, such as in the Coq
%   proof assistant, the user's code base is often significantly different from
%   the training data, containing definitions and theorems never seen during
%   training. We demonstrate two different methods of incorporating up-to-date
%   online information, each leading to state-of-the-art automatic theorem solvers
%   when tested on the challenging setting of proving theorems in Coq packages
%   never seen during training, while using only one CPU. First, an online $k$-NN
%   solver, based on the Coq Tactician, solves 25.8\% of theorems when given
%   online access to the most recent proofs in the code, compared to 15.0\% when
%   not. Second, we introduce a neural graph-based solver Graph2Tac, which uses a
%   novel approach to build hierarchical representations for new definitions not
%   seen during training. Graph2Tac's online definition task improves the theorems
%   solved from 17.5\% to 26.1\%. Moreover, the two online solvers are
%   complementary, together solving 33.2\% of theorems, outperfoming previous
%   state of the art approaches, including a transformer-based solver (14.8\%) and
%   the symbolic CoqHammer (17.4\%).
% \end{abstract}

\section{Introduction}

% - Explain the nature of formal math as a hierarchical structure.
% - Information that is closest in physical poximity is most important.
% - This is especially important for Coq, where new formalizations are often
%   started, with new domains
% - This goes contrary to the practice of offline training
% - We need the ability to incorporate new math concepts on-the-fly into a model.
% - We show the importance of learning online info through two models that use
%   orthogonal data, both of which are important.
%   + kNN learns from new proofs.
%   + Graph2Tac has a definition embedding task that learns to represent new
%   definitions.

Formal libraries of mathematics are structured hierarchically. Definitions build
on top of each other and lemmas are proven with the help of other lemmas. On a high
level, entire formalization projects concerning different mathematical topics
may also depend on one another. The result is an enormous web of formal
mathematical knowledge. In its center, one finds the most foundational parts of
mathematics such as logic and arithmetic. Towards the outskirts of the web, the
mathematics becomes increasingly specialized, as knowledge is built towards
specific goals such as proving the four-color theorem
\citep{10.1007/978-3-540-87827-8_28_fourcolor}. As the web of formal mathematics
matures, its core stabilizes into static knowledge while the outskirts are
ever-changing, as new insights emerge, connections are made and new theorems are
developed.

A good mathematician, whether human or artificial, should possess a deep and
intuitive understanding of the static, foundational core of mathematics. This is
gained through training and practice (in the case of an AI, on a large, offline
corpus). However, a mathematician must also have the capability to quickly gain
a working understanding of the concepts, lemmas, and proving techniques in a new
mathematical domain. This is a learned skill. An experienced mathematician can
grok a new math concept from its definition and perhaps by looking at a few
examples. For an AI, this amounts to ``online representation learning'': It must
have a procedure to quickly build new representations for concepts, without
having to retrain from scratch.

In this work, we explore online representation learning for the Coq proof
assistant \citep{the_coq_development_team_2023_8161141} through the Tactician
platform \citep{blaauwbroek_2023_10028721, DBLP:conf/mkm/BlaauwbroekUG20}. This
platform provides a large-scale dataset and machine-to-machine interaction
protocols that give an external proving agent (solver) online access to the full
internal formal knowledge base loaded into Coq. We evaluate the capabilities of
two online solvers, Tactician's built-in $k$-nearest neighbor ($k$-NN) solver
\citep{DBLP:conf/lpar/BlaauwbroekUG20, onlinerandomforest} and a novel model
called Graph2Tac against a number of baselines.

\begin{figure}[ht]
  \centering
  \includegraphics[scale=0.88]{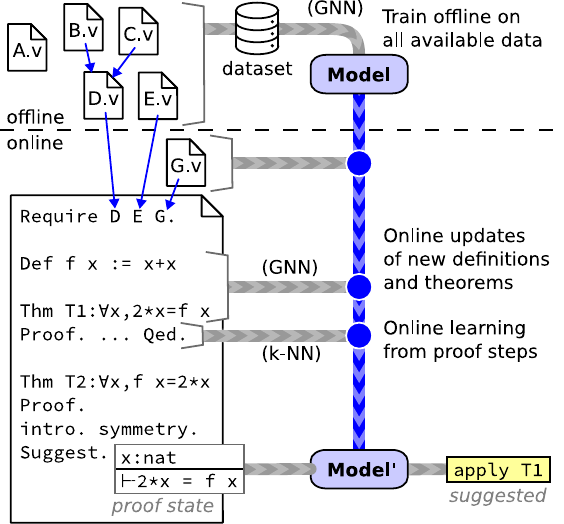}
  \caption{Our two paradigms of online learning: the online $k$-NN learns from recent proofs and the GNN (Graph2Tac) from recent definitions.}
  \label{fig:overview}
\end{figure}

Figure~\ref{fig:overview} summarizes the the paradigm
of both online solvers.
During offline training, the model
develops a deep understanding of the mathematics
found in a static Coq dataset.
It also learns to represent new information in an online manner.
During evaluation,
Graph2Tac incorporates information from new definitions and theorems,
and the online $k$-NN incorporates information from new proofs.
When the user invokes Tactician's \texttt{Suggest} command,
the model uses the current state to suggest a next proof step,
called a \emph{tactic} in Coq.
Each tactic is made up of an instruction followed by arguments.
When the user invokes Tactician's \texttt{synth} tactic,
a solver searches for a full proof of the current theorem,
calling the model at each step of the search.

The two online solvers we discuss in Section~\ref{sec:online-models}, $k$-NN and
Graph2Tac, operate on entirely different kinds of online data and as a result
have orthogonal capabilities (discussed in Section~\ref{sec:knn-g2t-comparison}).
The $k$-NN solver can learn on-the-fly from tactic scripts written by users.
As a result, it can make highly relevant suggestions when the proof of the
current lemma bears similarities to the surrounding proofs.

\begin{figure*}[t]
  \centering
  \includegraphics[scale=0.88]{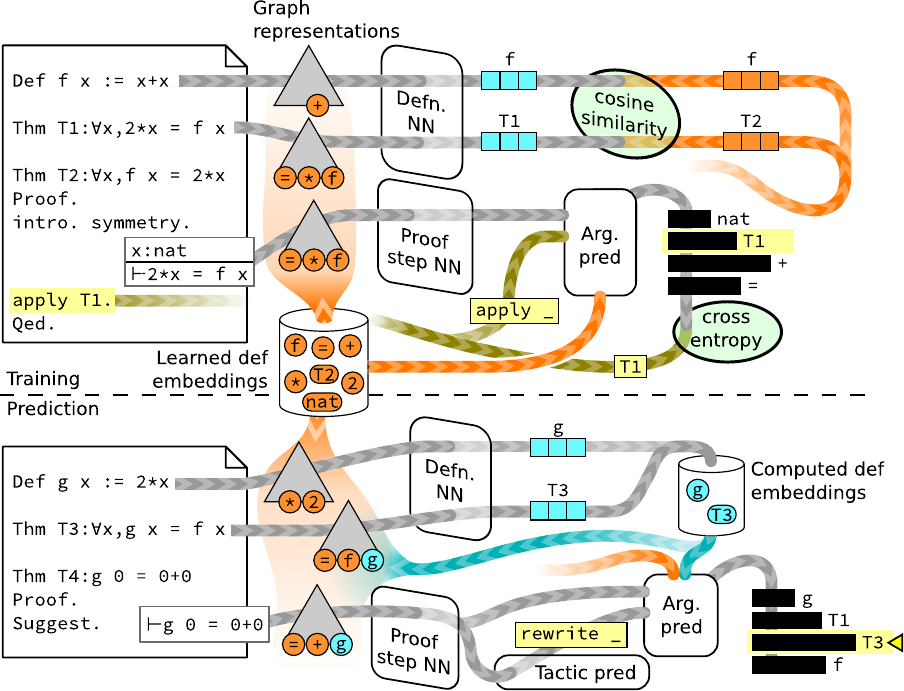}
  \caption{Our novel definition training task.
  At the center is a table of learned definition embeddings (orange).
  During training (above the line),
  we train both a proof step network to predict tactics (including arguments)
  and a definition network
  to calculate definition embeddings (blue) similar to those in the table.
  During inference (below the line),
  the definition network is used to calculated hierarchical embeddings (blue)
  for new definitions that may in turn be used to
  represent proof states and predict tactic arguments.
    }
  \label{fig:definition-training-task}
\end{figure*}

Graph2Tac, shown in Figure~\ref{fig:definition-training-task},
is a novel graph neural network (GNN) that,
in addition to standard offline learning capabilities,
contains a novel \textit{definition embedding neural network} and \emph{task}.
Our goal is to provide representations for definitions
(including theorems) both seen during training and inference.
For definitions seen during training,
the model learns a table of definition embeddings
(center of the figure).
These are used for representing definitions in proof states,
where the graph of each proof state is used by the proof step neural network
to suggest a tactic for that proof state.
They are also used for representing lemmas and other possible tactic arguments,
and are incorporated into argument prediction.
For new definitions, not seen during training,
the definition embedding neural network
computes an embedding from the definition graph.
In order to make sure these calculated embeddings
agree with the learned embeddings,
we co-train with a definition embedding task (upper right of the figure)
which keeps them aligned.
Each calculated embedding is hierarchical,
depending not only on the graph of this definition,
but the representations of component definitions.

\paragraph{Contributions}
We demonstrate the crucial importance of incorporating online information into
models. Comparing an offline and online version of the $k$-NN solver
demonstrates a proving performance improvement from 15.0\% to 25.8\% (1.72x).
Section~\ref{sec:gnn-def-task} introduces a new definition training task for
GNNs that improves performance from 17.4\% to 26.1\% (1.5x). Our Graph2Tac model
is based on an encoding of Coq's mathematical knowledge as a single
interconnected mono-graph (Section~\ref{sec:graph-based-interaction}) that
enables calculating hierarchical representations of definitions. Due to
integration with the Tactician platform, Graph2Tac is the first neural network
model conveniently available to Coq end-users. Graph2Tac and $k$-NN are highly
complementary, together proving 33.2\% of the test set.

We compare the online models to a variety of other general-purpose solvers for
Coq, including a transformer baseline, CoqHammer, Proverbot9001, ASTactic,
TacTok and Passport (Section~\ref{sec:non-graph}). Both online solvers prove at
least 1.48x more theorems than any of the comparisons, making them the strongest
by some margin. An extensive analysis of our experiments is in
Section~\ref{sec:evaluation} and several appendices. Our code is publicly
available, and every step of our work is fully reproducible
(Appendix~\ref{sec:reproducibility}).

\paragraph{Background and Related work}
This work fits into a growing body of work on machine learning and theorem proving.
For a full account, see Appendix~\ref{sec:extendedbackground}.
While the online capabilities of existing models are fairly limited,
one common source of online information is argument selection
(also called premise selection or lemma selection),
where the model selects useful lemmas or definitions arguments
for either the current theorem, proof state, or tactic.
In argument selection,
the proof state and lemmas are encoded via similar methods
and the lemmas most related to the proof states are selected.
Proof states and lemmas can be represented by
hand-crafted features \citep{LPAR-21:TacticToe_Learning_to_Reason},
S-expressions \citep{DBLP:conf/icml/BansalLRSW19, proverbot9001},
trees \citep{yang2019coqgym},
pretty printed text \citep{yang2023leandojo},
or graphs \citep{DBLP:conf/aaai/PaliwalLRBS20}.
They are then encoded with appropriate machine learning models.
Transformer-language based approaches \citep{han2021proof}
even forgo argument selection,
synthesizing lemma arguments directly from memory of the training data.
While convenient, this removes the ability to predict online arguments.

Tactician \citep{DBLP:conf/mkm/BlaauwbroekUG20}
uses a different online approach,
relying on online classifiers (including the aforementioned online $k$-NN)
to select tactic commands from available proof data.

However, none of these methods incorporate online definition information,
in neither the representations of proof states nor lemmas.
While some models may use the definition name either as a token or a string \citep{passport},
they do not ``look inside'' the definition.
This is the main novelty of Graph2Tac's hierarchical definition representation.

\section{Graph-based Interaction with Coq}\label{sec:graph-based-interaction}
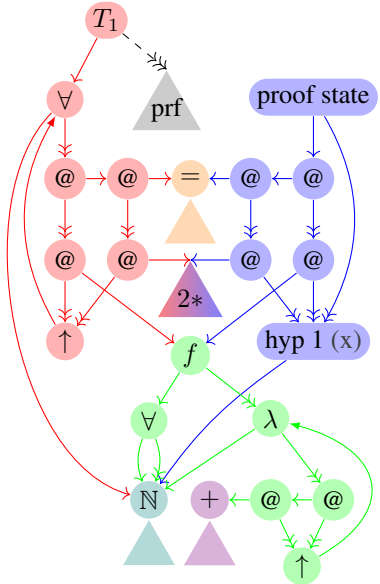
\begin{figure}[ht]
    \centering
    \hspace*{-10pt}
\begin{tikzpicture}[
  lnode/.style = {fill=gray!40, circle, inner sep = 1.5, minimum height=14, },
  nred/.style = {fill=red!30},
  nblue/.style = {fill=blue!30},
  ngreen/.style = {fill=green!30},
  npurple/.style = {fill=teal!30},
  nyellow/.style = {fill=violet!30},
  norange/.style = {fill=orange!30},
  mred/.style = {draw=red},
  mblue/.style = {draw=blue},
  mgreen/.style = {draw=green},
  mpurple/.style = {draw=pink},
  myellow/.style = {draw=violet},
  morange/.style = {draw=orange},
  ]
  \node[lnode, nred] (forall1) at (0, 0) {$\forall$};
  \node[lnode, nred, below = 0.8 of forall1, anchor=center] (app1) {@};
  \node[lnode, nred, right = 0.55 of app1, anchor=center] (app2) {@};
  \node[lnode, norange, right = 0.55 of app2, anchor=center] (eq) {$=$};
  \node[lnode, norange, regular polygon, regular polygon sides=3, inner sep = 4,
  below = 0 of eq, anchor = north] (eqt) {};
  \node[lnode, nred, below = 0.8 of app2, anchor=center] (app3) {@};
  \node[lnode, fill=white, regular polygon, regular polygon sides=3, inner sep = 0,
  right = 0.55 of app3, anchor = north,
  shade, left color=red!50, right color=blue!50] (2times1) {$2*$};
  \node[lnode, nred, below = 0.8 of app1, anchor=center] (app6) {@};
  \node[lnode, nblue, right = 0.55 of eq, anchor=center] (app7) {@};
  \node[lnode, nblue, below = 0.8 of app7, anchor=center] (app12) {@};
  \node[lnode, nblue, right = 0.55 of app7, anchor=center] (app11) {@};
  \node[lnode, nblue, below = 0.8 of app11, anchor=center] (app8) {@};
  \node[lnode, ngreen, below = 2.1 of eq, anchor=center] (f) {$f$};
  \node[lnode, nred, below left = 0.8 and 0 of app6.south, anchor=center] (v1) {$\uparrow$};
  \node[lnode, nblue, rounded rectangle, below right = 0.8 and 0 of app8.south, anchor=center] (hyp1) {hyp 1 \color{darkgray}(x)};
  \node[lnode, nblue, rounded rectangle, above = 0.8 of app11, anchor=center] (ps) {proof state};
  \node[lnode, ngreen, below left = 0.6 and 0.55 of f.south, anchor=center] (forall2) {$\forall$};
  \node[lnode, npurple, below = 0.8 of forall2, anchor=center] (nat) {$\mathbb{N}$};
  \node[lnode, npurple, regular polygon, regular polygon sides=3, inner sep = 4,
  below = 0 of nat, anchor = north] (natt) {};
  \node[lnode, nyellow, right = 0.55 of nat, anchor=center] (plus) {$+$};
  \node[lnode, nyellow, regular polygon, regular polygon sides=3, inner sep = 4,
  below = 0 of plus, anchor = north] (plust) {};
  \node[lnode, ngreen, right = 0.55 of plus, anchor=center] (app9) {@};
  \node[lnode, ngreen, right = 0.55 of app9, anchor=center] (app10) {@};
  \node[lnode, ngreen, above = 0.8 of app9, anchor=center] (lambda1) {$\lambda$};
  \node[lnode, ngreen, below = 0.6 of $(app9.south)!0.5!(app10.south)$, anchor=center] (v2) {$\uparrow$};
  \node[lnode, nred, rounded rectangle, above right = 0.8 and 0.55 of forall1.north, anchor=center] (T1) {$T_1$};
  \node[lnode, regular polygon, regular polygon sides=3, inner sep = -0.5,
  below right = 0.4 and 0.8 of T1.south, anchor = north] (proof) {prf};

  \draw[->>, mred] (forall1) -- (app1);
  \draw[->, mred] (app1) -- (app2);
  \draw[->, mred] (app2) -- (eq);
  \draw[->>, mred] (app2) -- (app3);
  \draw[->, mred] (app3) -- (2times1.north);
  \draw[->>, mred] (app3) -- (v1);
  \draw[->>, mred] (app1) -- (app6);
  \draw[->>, mred] (app6) -- (v1);
  \draw[-latex, mred] (v1) to[out=120, in=-120] (forall1);
  \draw[->, mred] (forall1) to[out=-135, in=165, looseness=0.8] (nat);
  \draw[->, mred] (T1) -- (forall1);
  \draw[->, mred] (app6) -- (f);
  \draw[->>, mblue] (app11) -- (app8);
  \draw[->, mblue] (app8) -- (f);
  \draw[->>, mblue] (app8) -- (hyp1);
  \draw[->, mblue] (app7) -- (eq);
  \draw[->, mblue] (ps) -- (app11);
  \draw[->, mblue] (app11) -- (app7);
  \draw[->>, mblue] (app7) -- (app12);
  \draw[->, mblue] (app12) -- (2times1.north);
  \draw[->>, mblue] (app12) -- (hyp1);
  \draw[->>, mblue] (ps) to[out=-60, in=60] (hyp1);
  \draw[->, mblue] (hyp1) to[out=-145, in=55] (nat);
  \draw[->, mgreen] (f) -- (forall2);
  \draw[->>, mgreen] (f) -- (lambda1);
  \draw[->, mgreen] (forall2) to[out=-110, in=110] (nat);
  \draw[->>, mgreen] (forall2) to[out=-70, in=70] (nat);
  \draw[->, mgreen] (lambda1) -- (nat);
  \draw[->>, mgreen] (lambda1) -- (app10);
  \draw[->, mgreen] (app10) -- (app9);
  \draw[->, mgreen] (app9) -- (plus);
  \draw[->>, mgreen] (app9) -- (v2);
  \draw[->>, mgreen] (app10) -- (v2);
  \draw[-latex, mgreen] (v2) to[out=30, in=-15, looseness=1.8] (lambda1);
  \draw[->>>, dashed] (T1) -- (proof.north);
\end{tikzpicture}
\caption{A graph-based representation of the example given in Figure~\ref{fig:overview}.
Each color represents a unique definition or proofstate subgraph.
These colored subgraphs are used as inputs to our neural networks.
Green corresponds to the function $f$,
red to the theorem $T1$, and
blue to the proof state from the example.
Each node is either a definition node ($T1$, $f$, $\mathbb{N}$, $=$, $+$) or
a node representing the part of a term.
For example, $@$ is function application,
and $\uparrow$ is a variable whose outgoing edge points to its binder.
Edges are labeled, but not shown here.
Triangles represent portions of the graph which we didn't expand for clarity.
Some nodes are shared by multiple colored regions such as the subgraph for $2*$.
Definition nodes (but not the rest of the definition)
are also shared with all colored regions they touch.
The proof terms for $T1$ is excluded from its colored region.
See \citet{web-paper} for full details.}
\label{fig:graph-example}
\end{figure}

In this work, we use a new graph-based data format to communicate between
solvers and Coq. In this format, all Coq definitions and proof states are stored
in one large interconnected graph~\citep{web-paper, blaauwbroek_2023_10028721}.
Figure~\ref{fig:graph-example} shows a subset of this large graph that
represents the proof state and environment at the start of the \verb|Suggest|
command in Figure~\ref{fig:overview}. This novel graph encoding provides the
following advantages: (1) The graph is designed to faithfully encode all kernel
objects in Coq's environment in a reversible manner (2)
References to global definitions, such as \verb|+| in Figure~\ref{fig:graph-example}, are
explicitly referenced by edges to the definition node,
avoiding any ambiguity with name resolution. (3)
Local variables are connected to their binders, \emph{e.g.}\ $\forall$,
$\lambda$, via edges. This eliminates the need for local variable names. (4) By
using a novel graph hashing algorithm~\citep{sharing-paper}, equal terms are
shared across parts of the graph leading to more compact representations.

This mono-graph can contain hundreds of millions of nodes and is
too large for a graph neural network. See Appendix~\ref{sec:big-graph} for an
illustration. Instead, our models hierarchically process subgraphs representing
proof states and definitions. Each proof
state and definition has a root node in the large graph. To obtain a smaller
graph, we calculate the forward closure from the root node, stopping when a
definition node is encountered. Theorems are special cases of definitions where
the root node references both the theorem statement and its proof term. We omit
the proof term from the theorem's definition graph to reduce the graph
size.\footnote{This is justified by the principle of \emph{proof irrelevance}:
  To use a theorem one does not need to know its proof.} In
Figure~\ref{fig:graph-example} each subgraphs received a different color. Each
color corresponds to either $T_1$, $=$, $f$, $+$, $\mathbb{N}$, or the current
proof state of theorem $T_2$, which is still under construction. Notice that
$T_1$ and the proof state share the subterm associated with $2*$ (not drawn in
full detail).

Mutually recursive definitions, such as inductive datatypes, have multiple root nodes
that reference each other. Such \textit{definition clusters} are treated as a
single unit. The subgraphs extracted from
the mono-graph are topologically ordered according to their dependencies, so
that they can be processed in an appropriate order by the neural network.

To train the model, a large dataset of proofs is extracted where each proof
state has an associated tactic. A tactic such as \verb|rewrite plus_comm in H|
is decomposed into a \emph{base tactic} \verb|rewrite _ in _|, and arguments
\verb|plus_comm| and \verb|H|. While arguments may be arbitrary Coq terms and
even other kinds of objects, our prediction model only predicts local hypotheses
or global definitions. See Appendix~\ref{sec:big-graph} for more information on
the dataset.

The Tactician \verb|synth| tactic and \verb|Suggest|
command can communicate via this graph format
with a Python client running a machine learning model.
Tactician sends the entire mono-graph of global definitions
along with the current proof state.
The client returns a list of suggested tactics and scores.
This integration makes our solver usable in practice
and allows us to perform a massively parallel benchmark of our model
on any Coq Opam package.

\section{Description of Solvers}
\label{sec:online-models}

Here, we describe the solvers that are part of the comparison in this paper,
including the online $k$-NN in Section~\ref{sec:knn-description}, our novel
Graph2Tac architecture~\ref{sec:gnn-def-task} and a number of baseline
comparisons in Section~\ref{sec:non-graph}.

\subsection{Online $k$-Nearest Neighbor}
\label{sec:knn-description}
The fastest practical solver currently available in the Tactician framework is
an online $k$-nearest neighbor ($k$-NN)
model~\citep{DBLP:conf/lpar/BlaauwbroekUG20}. It builds a database of proof
states and associated tactics and extracts hand-crafted features from those
proof states~\citep{onlinerandomforest}. When the model is queried, it looks up
proof states with the most similar features to the current proof state and
returns the corresponding tactics, ordered by similarity. It does not require
any training. The online $k$-NN uses a simple search over the 1000 most recent
tactic proof steps in the global context. We also compare to an offline version
that searches over all currently loaded proofs that are part of our training
data. The offline $k$-NN uses locality sensitive hashing forests (LSHF) to speed
up the search, similar to the $k$-NN solver found in Tactician\footnote{We do
  not limit the offline $k$-NN to 1000 examples because the obvious ``recency'' selection
  criterion does not apply.}. See
Appendix~\ref{sec:kNN-ablation} for additional $k$-NN variations and ablations.

\subsection{Graph Neural Network and Definition Task}\label{sec:gnn-def-task}

\begin{figure*}[t]
    \centering
    \includegraphics[scale=0.88]{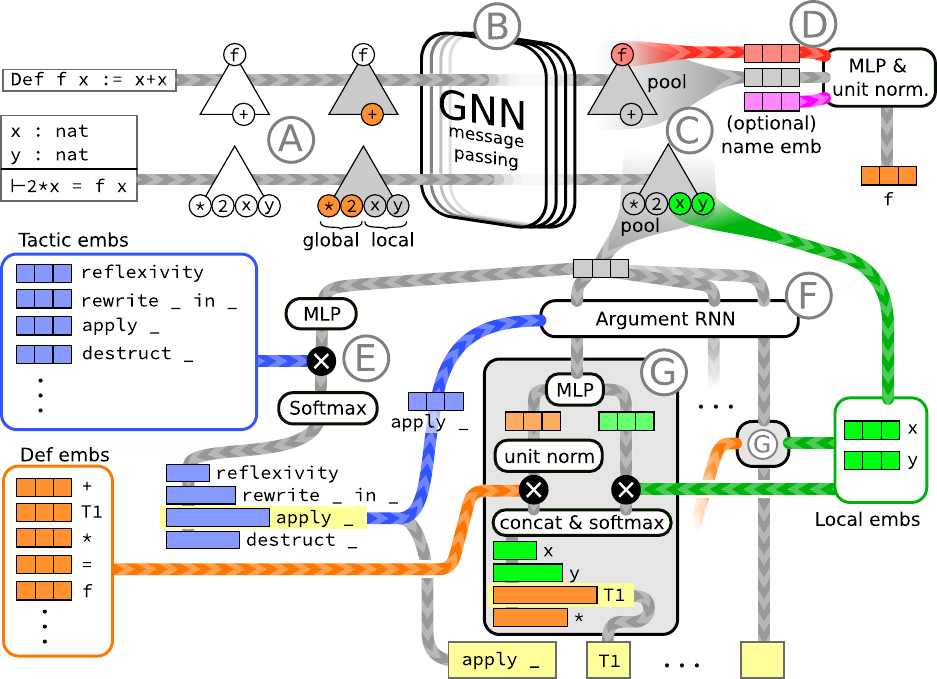}
    \caption{
    Detailed Graph2Tac model architecture,
    composed of a definition task
    (the top path going from the definition of $f$ to its embedding)
    and prediction task
    (the path going from the proof state $2 * x = f x$
    to the predicted tactic \texttt{apply T1}).
    The edge and node embedding tables are excluded from the image.}
    \label{fig:neural_network}
\end{figure*}

Graph2Tac primarily consists of two parts, a \emph{definition task} and a
\emph{prediction task}, shown in Figure~\ref{fig:neural_network}. The input to
each of these is a directed graph with labeled nodes and edges representing
either a definition for the definition task or a proof state for the prediction task.
We additionally associate metadata with each graph.
Definition graphs have \textit{root nodes} (node \verb|f| in the figure),
one for each definition defined in that graph.
Inductive data types define multiple mutually dependent concepts in the same graph,
\emph{e.g.} \verb|bool|, \verb|true|, \verb|false|,
and may therefore have multiple roots.
A proof state graph comes with \textit{local context nodes},
one for each hypothesis.
Both types of graphs contain \textit{definition nodes}, which are leaf nodes that
reference definitions in the global context.
With each definition node, we store an index to the definition embedding table
described in the next paragraph\todo{can this forward reference be resolved? Lasse}.
For the prediction task, in addition to the proof state graph,
the input includes the indices of all definitions and
tactics currently available in Coq's global context.

There are four learned embedding tables: edge labels and node labels (not shown in Figure~\ref{fig:neural_network}), base tactics that occur in the training dataset, and definitions that occur in the training dataset.  All embeddings have dimension equal to the model's hidden dimension, $h_\text{dim} = 128$.  The definition and node embeddings are constrained to be unit-normalized. During inference, the definition table will be dynamically updated with new definitions as discussed later.

We transform the input graph, be it a proof state or definition graph, using the
following approach (Figure~\ref{fig:neural_network}, step A).  Each graph is pruned to 1024 nodes.  For effective message passing, we duplicate all edges to go in both directions and also add self-edges.  The edges are assigned edge embeddings from the corresponding embeddings table.  This table has $2E+1$ entries, accounting for the $E$ original edge labels, the $E$ reverse edge labels, and the self-edge label.  Non-definition nodes are assigned embeddings from the node embeddings table based on their node label.  Definition nodes are instead assigned embeddings from the definition embeddings table, except for the root nodes in the definition task.  Those are masked with null embeddings, as the goal of the definition task is to predict the corresponding definition embeddings.

The transformed graphs are put into a message-passing GNN (Figure~\ref{fig:neural_network}, step B). The GNN consists of 8 hops where the $t$th hop transforms a node embedding $x^{t}_n$ according to the following two steps:
\[
\hat{x}^{t+1}_n = \text{ReLU}\left(\frac{1}{\text{deg}(n)}\sum_{m, e\colon m\xrightarrow{e} n} \text{Dense}_{\theta_t}(e, x^{t}_m)\right)
\]
\[
x^{t+1}_n = \text{Layernorm} \left(x_t + \text{Dropout}_{0.1}(\text{MLP}_{\psi_t}(\hat{x}^{t+1}_n)) \right)\]
The first step is a graph convolution layer where each node embedding is updated according to incoming edges from neighboring nodes.  Here, $\text{deg}(n)$ is the number of incoming edges $m\xrightarrow{e} n$ with target $n$ and edge embedding $e$.  The dense layer has output dimension $h_\text{dim}$.  The next step consists of a 2-layer MLP (Dense, ReLU, Dense) with an inner hidden dimension of $2h_\text{dim}$, then Dropout, Residual, and Layernorm.  The weights for each hop are separate, but the definition and prediction tasks both use the same GNN backbone, sharing the same weights.

The output of the GNN is a graph with the same structure as the input, but with updated node embeddings. Both the definition and prediction tasks use mean pooling (Figure~\ref{fig:neural_network}, step C) to obtain a single vector embedding for the graph.
For the definition task, the pooled embedding is concatenated with each of the embeddings of the root nodes for the definition, then fed into a two-layer MLP, and finally unit normalized (step D).
Optionally, along with each root node embedding,
we additionally concatenate a name embedding for the definition,
using a bidirectional LSTM (not-shown) to embed the fully
qualified Coq identifier string, \emph{e.g.}\ ``Coq.Init.Logic.and''.
For the prediction task, the pooled embedding is fed into a two-layer MLP (step
E).  The output is multiplied by the tactic embedding table using inner product
and put through a softmax layer to obtain the base tactic probabilities.

To predict the arguments for a tactic, we use a simple two-layer RNN (Figure~\ref{fig:neural_network}, step F).  The initial hidden state of the RNN is the embedding for the chosen tactic in the tactic embedding table.  (During training the ground truth tactic is provided as input.  The decoding method used during inference is described below.)  The number of steps of the RNN corresponds to the number of arguments required by the given tactic.

Each argument is predicted as follows (Figure~\ref{fig:neural_network}, step G).
The RNN output is fed into a pair of two-layer MLPs (Dense, ReLU, Dense),
resulting in a pair of query embeddings,
one for global argument prediction and one for local argument prediction.
The global argument prediction query is unit normalized.
For global arguments, we take the inner product of the global argument query
with each definition in the embedding table
resulting in a logit for each definition.
Since the inner product is bounded
(as both vectors in the inner product are unit normalized),
we scale the global logits by a learned temperature parameter.
For each local argument,
we use the GNN output embedding for that local node in the graph
and take the inner product with the local argument query.
Concatenating the local and global logits
and performing softmax we get a distribution over all possible arguments.

Since many of the calculations used in our model take place on
variable-size lists of objects, \emph{e.g.} local arguments,
our implementations rely heavily on
ragged tensors in TensorFlow~\citep{tensorflow2015-whitepaper}
and graph tensors in TF-GNN~\citep{tfgnn}. 

We train on batches of 512 definitions and 512 proof states.  The loss for the definition task is cosine similarity.\footnote{
If a definition graph contains multiple definitions,
the loss is divided by $\sqrt n$ where $n$
is the number of entry points.
}
The loss for the training task is the cross-entropy for the full tactic sequence.  For example, for the tactic $\verb|apply T1|$ the loss is $-\log P(\verb|apply _|) - \log P(\verb|T1| \mid \verb|apply _|)$ using the probabilities coming from the model.  The combined loss of the two tasks is
$\mathcal{L} = 1000\, \mathcal{L}_{\mathit{def}} + \mathcal{L}_{\mathit{tactic}}$.
% Note to self.  1000 is 500 in our code since I'm talking here about cosine similarity and not l2 norm (which has a derivative of twice the cosine similarity).

During inference, the tactic embedding table is masked for tactics which both
occur in the training data and are available in the Coq's current state.
Similarly, the definition embedding table is masked for all definitions
available in the current global context.  However, for new definitions not seen during training, we first calculate an embedding using the definition task.  If there are multiple new definitions, we compute embeddings from each definition graph individually, updating the embeddings in a topologically sorted order so that those for dependencies are computed before those for latter definitions which depend on those dependencies.

At inference time, the output of the tactic task is a list of tactic suggestions, where each sequence starts with a base tactic, \emph{e.g.}\ \verb|apply _| and then contains the arguments for that tactic, if they are required. We use beam search decoding with a beam width of 256 to generate 256 tactic suggestions for each proof state.

We train three models:
G2T-Named uses the name embedding in step D, whereas G2T-Anon does not.
G2T-NoDef is trained without a definition task.
Each of these is run with three configurations on the definition model during inference:
The Recalc configuration calculates embeddings for \emph{all} definitions,
Update only calculates embeddings for \emph{new} definitions---\emph{i.e.}\
those not seen during training,
and Frozen uses random unit normalized embeddings in place of the definition model.
G2T-NoDef is only used as G2T-NoDef-Frozen.
G2T-Anon-Update is the primary model.

\subsection{Baselines for Comparison}\label{sec:non-graph}
We formally compare the Graph2Tac and $k$-NN models with three baselines
outlined below. Note that some obvious baselines, such as
SMTCoq~\citep{DBLP:conf/cpp/ArmandFGKTW11} and
Itauto~\citep{DBLP:conf/itp/Besson21} were excluded because they are not general
purpose solvers. Furthermore, the online random forest described by
\citet{onlinerandomforest} is excluded because it does not scale to large
formalizations.

\textbf{\texttt{firstorder auto with *}}
As a baseline, we use Coq's built-in \texttt{firstorder} reasoning tactic
combined with the programmable proof search tactic \texttt{auto}, with all
available hint
databases.

\textbf{CoqHammer}~\citep{coqhammer} translates theories expressed in Coq to a first-order language
understood by the external automated theorem provers Vampire, CVC4, Z3 and
Eprover. Once an external proof is found, the premises required for the
proof are extracted, and the proof is reconstructed inside of Coq through the
\texttt{sauto} family of higher-order solving tactics~\citep{sauto}.  \todo{Check typographical conventions (italics for Coq packages, monospaced fonts for code or commands, etc)}

\textbf{Proof State Text-based Transformer}
We implement a decoder-only transformer baseline that operates on the textual representations of the proof states,
and predicts a textual representation of the tactic.
We use the GPT2 implementation available via the Transformers
Python library~\citep{wolf-etal-2020-transformers}. The embedding size is set to
768 and it has 12 layers, as in one of the models described in that paper.
Our approach is similar to that of~\citet{han2021proof,DBLP:conf/nips/JiangLTCOMWJ22,yang2023leandojo}, except our transformer models are
trained from scratch only on Coq proof data.

\textbf{Informal Comparisons}
Appendix~\ref{sec:CoqGym} contains informal comparisons to highly relevant
neural solvers such as Proverbot9001~\citep{proverbot9001},
ASTactic~\citep{yang2019coqgym} and their derivatives. Due to technical
differences, a direct comparison was not achievable.
Nonetheless, we are
confident that Graph2Tac and $k$-NN significantly outperform these solvers for
two reasons. First, CoqHammer is known to outperform these neural solvers. Given that
Graph2Tac and the $k$-NN significantly outperform CoqHammer, we can conclude that
they also surpass the other neural solvers. Second, Appendix~\ref{sec:CoqGym}
contains informal experiments that directly compare to the $k$-NN solver.

\subsection{Capabilities of Graph2Tac vs $k$-NN vs Transformer}
\label{sec:knn-g2t-comparison}

\begin{figure*}
    \centering
    \includegraphics[width=1.0\textwidth]{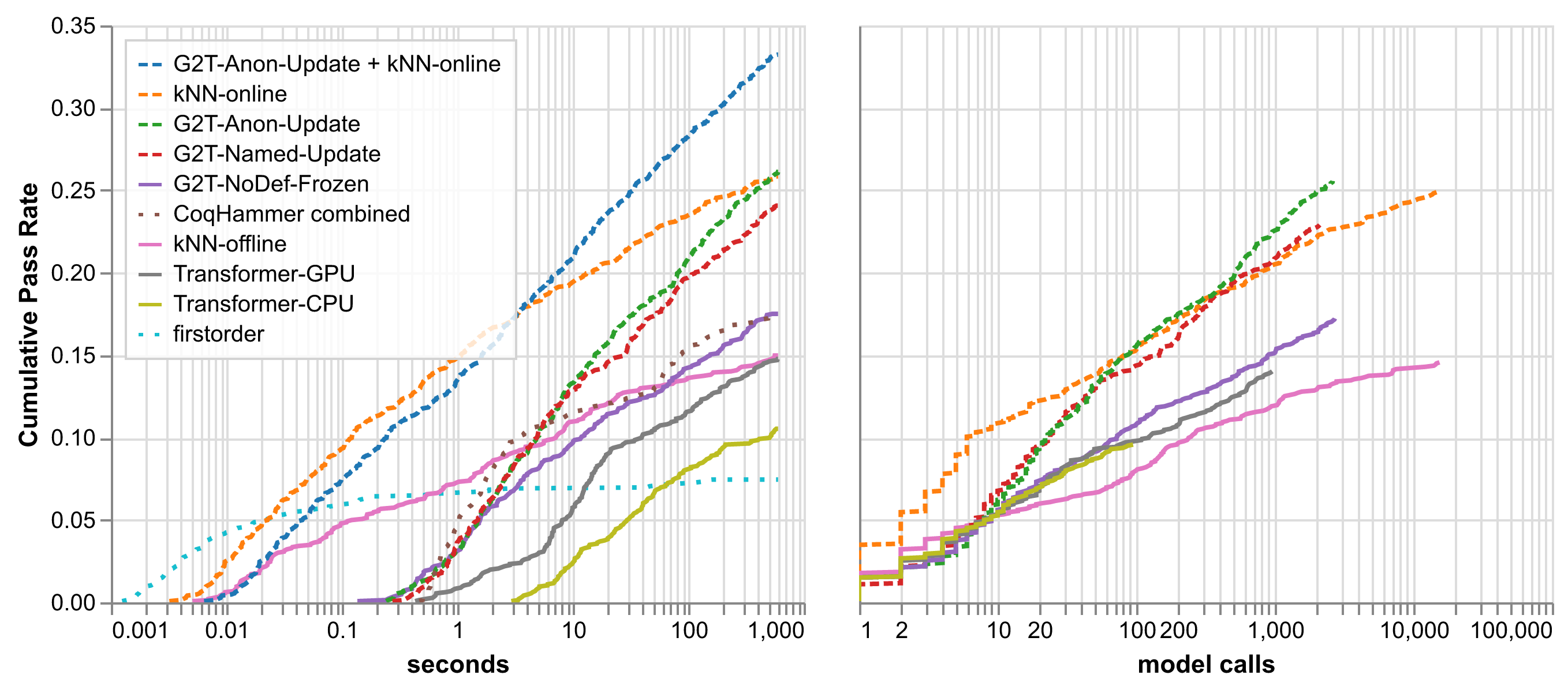}
    \vskip -0.1in
    \caption{Fraction of theorems proved over time and per number of calls to the model.  Online solvers are dashed, offline solvers are solid, and symbolic solvers are dotted.}
    \label{fig:test-packages-time}
\end{figure*}
The Graph2Tac, online $k$-NN, and transformer solvers are all based on the
Tactician platform. Nevertheless, their input and output capabilities are quite
different. The online $k$-NN model sees tactics and their arguments as a black
box. It can only predict the {\em exact} tactic expressions available
in its database. Its power comes from speed and the ability to amend its
database with new proof scripts. Graph2Tac uses a fixed set of tactics and
cannot learn from new proof scripts. However, it can choose lemmas and
hypotheses as arguments for these tactics. Due to its
definition embedding task, it can predict lemmas not available in the training
set. Finally, the transformer has no online learning capabilities. However,
unlike the $k$-NN and Graph2Tac, it can generate tactics with arbitrary terms as
arguments (but it must ``hallucinate'' references to new lemmas).

\section{Evaluation}\label{sec:evaluation}
\paragraph{Experimental Setup}
To evaluate the performance of the solvers described above, we randomly choose
2000 theorems from the Coq packages in our testing set. The solvers are given a
time limit of 10 minutes per theorem to synthesize a proof.
To look at package-specific results,
we sample up to 500 test theorems per package with a time limit of 5 minutes.
The search procedure utilized by the solvers based on the Tactician
framework is a modification of Dijkstra's algorithm that performs iterative
deepening in order to relieve memory pressure and has an optimization that
avoids backtracking between independent subgoals. For more information, see
Appendix~\ref{sec:search-procedure}.

During evaluation, each solver is limited by the operating system to one CPU
with two hyperthreads. All processes, including Coq and any external processes
such as neural networks and ATP's, must share this CPU. An exception is made for
the Transformer-GPU solver, which is permitted to utilize a dedicated GPU.

We explore two artificially aggregated solvers, that simulate running $n$
solvers concurrently while still adhering to the one-CPU computational budget.
The number of theorems ``solved'' by the aggregate solvers in time $t$
is the number solved by any of the components in time $t/n$.
``G2T-Anon-Update + $k$-NN-online'' is intended to simulate a solver capable of
learning both from new definitions in the global context and new tactic proof
scripts.
``CoqHammer combined'' is an aggregate of all four ATP backends and the
reconstruction tactic called \texttt{best}. This simulates the intent of
CoqHammer's authors to run many backends in parallel, while maintaining a fair
comparison with other solvers.

\paragraph{Results}
\begin{figure}[h]
  \centering
  \includegraphics[width=0.8\columnwidth]{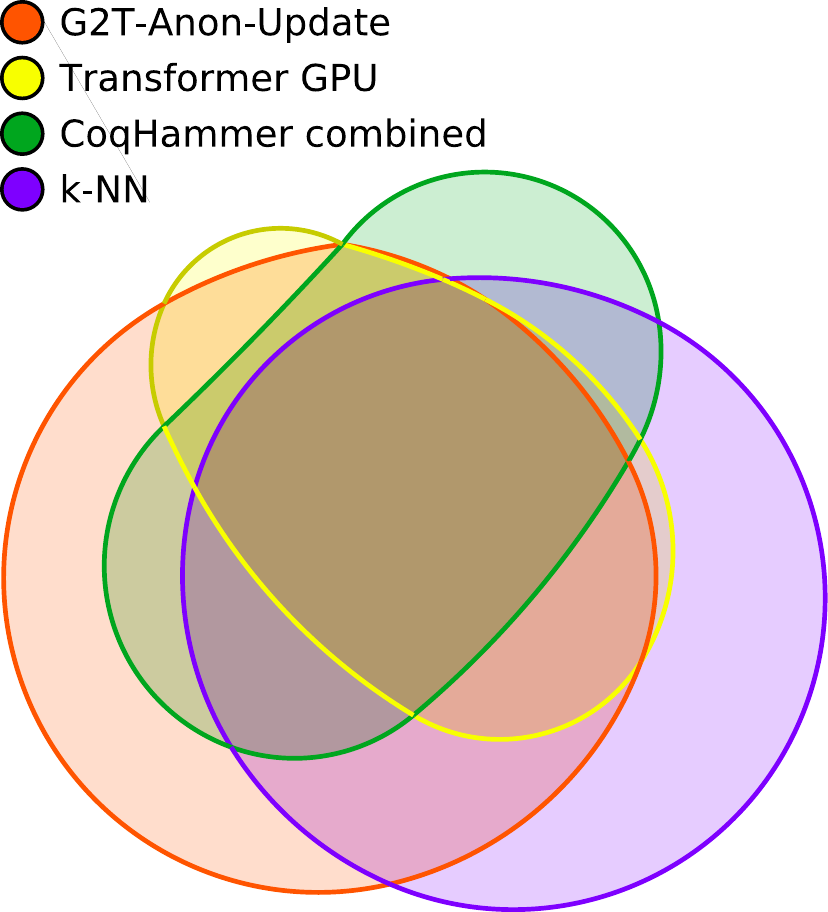}
  \caption{Venn diagram of the theorems proved. }
  \label{fig:venndiagram}
  \vskip -0.2in
\end{figure}

The left plot of Figure~\ref{fig:test-packages-time}
shows the fraction of test theorems solved over time for various solvers.
One can see both the progress over time
and an indication of the startup cost of each solver.
The right plot of Figure~\ref{fig:test-packages-time}
replaces time with the number of calls to the underlying model,
giving an indication of the usefulness of that model's predictions
in search irrespective of model speed.\footnote{
Pass rate is also impacted by the number of
tactic suggestions returned in a model call.
Too few tactics may lead to the search terminating early.
Our faster models also return more suggestions.
}.  This and other plots remove the results from the \textit{hott} package
because HoTT replaces Coq's built-in standard library,
upon which CoqHammer depends.
The \textit{tlc} package is also removed because G2T-Anon-Update
was able to exploit an inconsistency as explained in
Appendix~\ref{sec:axioms}.

Among the $k$-NN solvers, the online $k$-NN solves
many more theorems (25.8\% vs 15.0\%) showcasing
the importance of incorporating of recent online proof data.
(Further $k$-NN ablations are found in Appendix~\ref{sec:kNN-ablation}.)

Among the G2T solvers, the online solvers using the definition task
to compute embeddings for new definitions
outperform the G2T-NoDef-Frozen baseline
(26.1\% vs 17.4\%)
which does not compute embeddings for new definitions.
Note our baseline G2T-NoDef-Frozen performs similarly
to the transformer baselines
in terms of model calls (and even better in per-time performance).
This suggests that our baseline is at least comparable to the
common transformer baselines in other papers.
The addition of names in G2T-Named-Update fares slightly worse
than the main G2T solver G2T-Anon-Update.
(Further G2T ablations are found in Appendix~\ref{sec:g2t-ablation}.)

CoqHammer combined fairs better than the transformer solver,
even the variant using a GPU for model inference.
See Appendix~\ref{section:CoqHammer}
for a detailed breakdown of the CoqHammer results.

The $k$-NN solvers
benefit from being fast,
hence both $k$-NN solvers excel at lower time limits.
In Appendix~\ref{sec:CoqGym} we suggest the online $k$-NN is at least as
powerful as existing neural Coq solvers in the literature.
However, for higher time limits,
the G2T-Anon-Update model starts to overtake,
suggesting that if larger amounts of search are required,
there is value in more sophisticated neural models.
It is similar relative to model calls.

Moreover, the online $k$-NN and G2T approaches
are quite complementary as 
the combined solver G2T-Anon-Update + $k$-NN
is much better than either approach alone.
We also see this in the Venn Diagram in Figure~\ref{fig:venndiagram}.

\begin{figure}[h!]
    \centering
    \includegraphics[width=1.0\columnwidth]{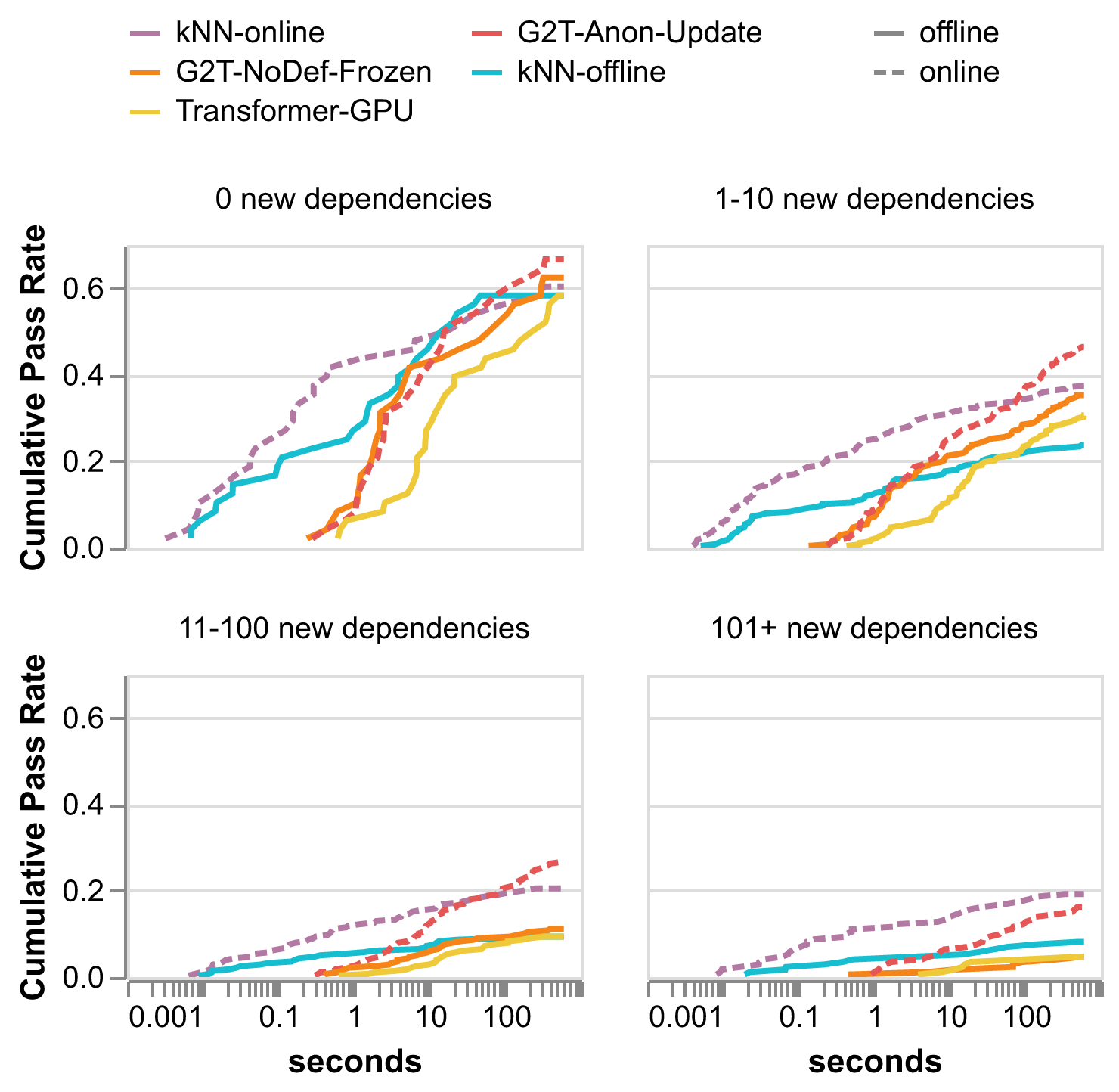}
    \vskip -0.1in
    \caption{Cumulative pass rate by number of new dependencies.}
    \label{fig:dependency_results}
\end{figure}

To investigate the effect of the online setting,
Figure~\ref{fig:dependency_results}
breaks down per-time pass rate by the number of new dependencies
not seen during training, including transitive dependencies.
The online models significantly outperform the offline models
when a theorem has at least 10 new dependencies.
See Appendix~\ref{sec:online-setting} for a more in-depth analysis.

%Figure~\ref{fig:package-performance-grid} breaks down the performance
%for the largest 15 packages.
%We see that neither the G2T-Anon-Update nor the $k$-NN always performs
%better.
%In many cases G2T-Anon-Update either overtakes the $k$-NN or appears as if it will overtake the $k$-NN if given enough time.
%The \emph{tlc} package is special in
%that G2T-Anon-Update was able to find a significant number of
%theorems using an inconsistent axiom in the environment,
%which is why it was removed from the results above.

%\begin{figure}[h!]
%    \centering
%    \includegraphics[width=1.0\textwidth]{images/grid_plot_with_ch_15_packages.png}
%    \caption{Package-specific cumulative solving curves. We show the behaviors of $k$-NN, GNN, and CoqHammer (except on HoTT which is incompatible with CoqHammer).}
%    \label{fig:package-performance-grid}
%\end{figure}

\section{Discussion and Future Work}
Our definition task improved a neural theorem prover from 17.4\% to 26.1\%
in the difficult setting of proving theorems in never-before-seen packages.
This, in addition to the success of the online $k$-NN approach,
shows the importance of online learning in this setting.
We leave as future work how to unify the G2T and online $k$-NN
approaches shown in Figure~\ref{fig:overview}.
Ideally, such a model should also account for new tactics,
as well as learn from how new definitions and tactics are used in new theorems.
One avenue is exploring if our model can be fine-tuned in real time.

To improve the definition task,
given that G2T-Anon-Update outperformed G2T-Named-Update,
we wonder if adding the names makes the definition task too easy for the model.
There may also be alternatives to our definition task,
using ideas in self-supervised or contrastive learning,
or using social-network-size graph models
to process the entire graph of interconnected definitions at once.

Theorem proving and programming share a lot of similarities and concerns, so it
is useful to explore how this work relates to code generation. We leave open how
our methodology could apply to text-based models. Retrieval augmented
transformers are a possible approach~\citep{yang2023leandojo}, but may not scale
to the full definition hierarchy.

\clearpage
\section*{Impact Statement}
In a world increasingly dependent on technology,
the need for verifiably safe and secure hardware and algorithms
is becoming increasingly urgent.
This is even more so as a larger amount of computer code is written by AI agents.
While interactive theorem provers, such as Coq,
have been repeatedly used to verify critical systems,
the process is still too labor-intensive for widespread adoption.
Machine learning research, such as ours, which improves the state of the art in
automated theorem proving for ITPs has the potential to revolutionize
the development of safe and secure code.

\section*{Acknowledgments}
This work was partially supported by the European Regional Development Fund
under the Czech project AI\&Reasoning no.~CZ.02.1.01/0.0/0.0/15\_003/0000466,
Amazon Research Awards, EU ICT-48 2020 project TAILOR no.\ 952215, and by the
Czech MEYS under the ERC CZ project POSTMAN no.~LL1902. Lasse Blaauwbroek
acknowledges travel support from the European Union’s Horizon 2020 research and
innovation programme under grant agreement No 951847.

We would also like to thank Alex Sanchez-Stern and Emily First for providing the data used in Appendix~\ref{sec:CoqGym}.

\bibliography{iclr2024_conference}
\bibliographystyle{iclr2024_conference}
\appendix

\section{Reproducibility}
\label{sec:reproducibility}
This paper carefully describes our methods for building, training, and testing
our models and solvers. Our solvers are intended for use by both researchers and
Coq end-users. For end-users, we provide optimized, pre-trained models that can
integrate directly with Coq. Installation and usage instructions can be found on
Tactician's website\footnote{Installation instructions:
  \url{https://coq-tactician.github.io/api/graph2tac}
}. The models run on a
typical Linux or x86 Mac machine and do not require a GPU. The model can also be
hosted on an external server (for example, with a GPU), and used by a locally
running Tactician instance, via a TCP connection.

In this paper, we use a freely available dataset of Coq
Data~\citep{blaauwbroek_2023_10028721}. Information on the construction of the
dataset can be found in \citet{web-paper} and \citet{sharing-paper}. The dataset
is freely available under the original licenses of its constituent Coq projects.
Most of these licenses are open source. However, some licenses place
restrictions on commercial usage. All our models are trained from scratch,
without the use of proprietary pre-training data.

To enable the full reproduction of the experiments in this paper, all code,
datasets, benchmark parameters, benchmark results, and analysis scripts are
available in a separate artifact~\citep{rute_2023_10410474}.
It includes, among others, the following elements:
\begin{itemize}
\item Archived versions of all code repositories used in the project. Among
  others, they include the following:
  \begin{itemize}
  \item Graph2Tac: Code to train and evaluate the GNN.\footnote{Github: 
      \url{https://github.com/IBM/graph2tac}
      \newline \null\quad\quad\!\! PyPI: \url{https://pypi.org/project/graph2tac}
    }.
  \item Text2Tac: Code to train and evaluate the Transformer.\footnote{Github: 
      \url{https://github.com/JellePiepenbrock/text2tac}
      \newline \null\quad\quad\!\! PyPi: \url{https://pypi.org/project/text2tac}
    }
  \item The main Tactician
    repository.\footnote{\url{https://github.com/coq-tactician/coq-tactician}
    }
  \item The code repository for Tactician's API used to integrate external
    agents into
    Coq.\footnote{\url{https://github.com/coq-tactician/coq-tactician-api}
    } This
    also includes the PyTactician Python library that can be used to extract
    data from the dataset and interface with Coq.\footnote{PyPi:
      \url{https://pypi.org/project/pytactician}
      \newline \null\quad\quad\!\!\! Docs: \url{https://coq-tactician.github.io/api/pytactician-pdoc}
    }
  \item The benchmarking system used to evaluate
    models.\footnote{Github: \url{https://github.com/coq-tactician/benchmark-system}
    } This
    system allows massively parallel evaluations of agents on all Coq projects
    packaged in the Opam package manager.
  \end{itemize}
\item Trained models for Graph2Tac and Text2Tac. They correspond exactly with
  the models used to obtain the results in this paper. We also include the
  scripts and parameters used to train these models.
\item Benchmark scripts that were used to invoke the benchmarking system for
  each agent evaluated in this paper. The scripts include the precise parameters
  used to invoke the models, as well as a reference to the precise Git commit of
  all involved code.
\item The raw data for the benchmarks used in this paper. It includes which
  theorems were tested on, which were proved, and how long the solver took in
  terms of seconds, model calls, and tactic executions. This is all in hopes of
  facilitating future comparison and collaboration.
\item Analysis notebooks that distill all the raw data into the curated data as
  presented in this paper. They will allow the interested reader to verify our
  results and perform further analysis.
\end{itemize}

The artifact should enable anyone to replicate each stage of our results,
including re-training the models, re-running all benchmarks, and re-doing
our analysis. Although reproducing the entire pipeline would require significant
work, we expect that our benchmarking data can be re-used easily for comparison
against future solvers.

Users can also train and evaluate their own agents with the code we provided and
the open dataset. While we trained the graph models for three weeks from scratch
on two A100s,
in Appendix~\ref{sec:training-time} we show that a model trained in two days achieves similar
results.  Users may also train models on a different set of Coq packages,
including new or private Coq developments, via the data extraction tools
provided with the dataset and with our training code.

\section{Threats to Validity}
Like other results in machine learning, there are many subtleties that
we may not have accounted for in this research,
but here we list the important points to look out for.

In theorem proving, it is a known challenge to compare to other comparable solvers.
We are unable to provide a direct comparison in our benchmarking system
to most existing Coq solvers, and the informal comparison in Appendix~\ref{sec:CoqGym}
is inconclusive due to methodological differences.
Greater challenges exist for comparing to solvers in HOL-Light, Isabelle, and Lean.
More standardized benchmarks are needed.

The comparisons we did perform also run the risk
of not being an apples-to-apples comparison.
The G2T and transformer models use different setups.
The former uses TensorFlow and custom-written beam search,
while the latter uses standard machine learning APIs for PyTorch
which may not have been as optimized for our setting.
Further, our baseline transformer solver is only an approximation of similar work.
The transformer models of~\citet{polu2020generative, han2021proof, DBLP:conf/nips/LampleLLRHLEM22, DBLP:conf/nips/JiangLTCOMWJ22, yang2023leandojo}
are larger, pre-trained, and use more resources.
It may be that emergent capabilities arise from
scaling which are not seen in our smaller models.
However, one reason we chose to avoid pretraining
(besides parity with our graph model)
is the strong risk that our test data will be leaked into the pretraining data.
(See the discussion on this topic in~\citet{han2021proof}.)
Also, we were specifically exploring models that could be
run with the resources available to a typical Coq user.
Similarly, our choice to run CoqHammer on a single thread with default settings
may not showcase its strongest capabilities.
See the discussion in Appendix~\ref{section:CoqHammer}.

The evaluation in Section~\ref{sec:evaluation}
and the further ablation in Appendix~\ref{sec:g2t-ablation}
of the G2T family of models relies on comparison with
the G2T-X-Frozen models using random embeddings.
There may be alternative ways to perform this baseline.
Passport~\citep{passport} uses definition embeddings from names.
HOList~\citep{DBLP:conf/icml/BansalLRSW19} and
ProverBot9001~\citep{proverbot9001} use a special token for unknown definitions.

Our choice to split on package was essential to testing our online setting,
and comparable to other Coq solvers,
but it also means our results are more variable.
A different split would likely give different pass rates.
Even our decision to exclude \emph{hott} and \emph{tlc} from our results
heavily influences our final results.
This is different from papers in other ITPs
which split randomly by theorem instead of package.
Similarly, we only train one of each model instead of multiple instances,
and our models were difficult to reliably train.
This also increases the variability of our results.

The data we use decomposes tactics into smaller component tactics.
This is in contrast to models like PACT~\citep{han2021proof} which
not only do not decompose tactics but also train on combinations of tactics.
It is possible our decomposition leads to more steps needed for a proof,
and this may especially punish slower models like the transformer
which can generate more complicated tactic combinations.
Also, our G2T model uses a fairly simple representation of tactics.
There are 100s of base tactics
(we excluded any that occurred fewer than 6 times)
and one Coq tactic, \emph{e.g.} \verb|apply|,
may be represented as many different base tactics in our model
(\emph{e.g.} one for each possible number of arguments).
Also, we only allow tactic arguments which
are single definitions or local context hypotheses.
The G2T model has no way to predict term arguments,
which are common in Coq proofs.

During benchmarking, it was common to encounter Coq errors.
In the end, any error we counted as a failed proof attempt.
Some systematic errors we fixed with patches to Coq,
but different solvers led to different errors,
and some were easier to fix than others.
Further, an error in a test theorem could even cause
an error in a later test theorem in the same file.
CoqHammer had a larger number of errors.

\section{Extended Related Work}
\label{sec:extendedbackground}
We take some extra space in this appendix to touch upon more related work. 

In recent years, there have been machine learning approaches for various interactive theorem proving systems. In the Coq ecosystem specifically, a series of articles \citep{tactok, diva,passport} was based on the dataset provided by CoqGym \citep{yang2019coqgym}, which contains many Coq packages to benchmark machine learning systems. Other projects in similar directions were GamePad~\citep{DBLP:conf/iclr/HuangDSS19}, an interaction system based on the coqtop REPL, and ProverBot9001~\citep{proverbot9001}. Another line of work is based on the Tactician system~\citep{DBLP:conf/mkm/BlaauwbroekUG20}, for which implementations of k-nearest neighbors and random forest algorithms were built \citep{onlinerandomforest}. Early systems for Coq were Proof General~\citep{proofgeneral} and SEPIA~\citep{SEPIA} and the experiments by \citet{DBLP:conf/sycss/KaliszykMU14}. A system that is often used for comparisons is CoqHammer \citep{coqhammer}.

For other interactive theorem proving systems, there has also been a lot of work done in recent years. Machine learning guidance was used to prove problems from the Mizar ITP system \citep{DBLP:conf/mkm/Urban03,kaliszyk2015mizar}. TacticToe~\citep{LPAR-21:TacticToe_Learning_to_Reason} is an ITP machine learning system for HOL4. For HOLight, there is the HOList~\citep{DBLP:conf/icml/BansalLRSW19} system. In Isabelle, the Sledgehammer system was also extended with machine learning \citep{DBLP:conf/itp/KuhlweinBKU13, DBLP:journals/jar/BlanchetteGKKU16}. For the recently developed Lean system, there are, for example, LeanDojo \citep{yang2023leandojo}, integrated decision trees
\citep{piotrowski2023machine}, and work using Language Models (LM) \citep{han2021proof,DBLP:conf/nips/LampleLLRHLEM22}. There have also been LM-based approaches for other systems, for example Magnushammer \citep{mikula2023magnushammer} for Isabelle and the GPT-f system~\citep{polu2020generative} for MetaMath.

Our work is novel compared to the previous related work: we use a graph-based dataset extracted with Coq kernel knowledge which allows us to develop a graph neural network that learns the meaning of definitions by exploiting the hierarchical structure of the data. This architecture tackles the problem of learning online, making use of new definitions in new projects. \todo{not sure if this paragraph is necessary, probably covered by other parts}

% In the CoqGym family of solvers [REF: CoqGym, TakTok, Diva], they use tree-based representations of proof states while we use graph based representations, but more importantly, their representations don’t use information about the definition nodes in their trees.  The one exception to this is [REF: Passport] which uses the names of definitions as a representation, while our approach uses the content of the definition itself.

% F
% ACL2ml\citep{acl2ml}

% We introduce our graph neural network architecture, Graph2Tac, which can naturally make use of new definitions. 

% Roosterize \citep{roosterize}
% This seems to be mostly concerned with suggesting lemma names?

% Proofster \citep{proofster}

In interactive theorem proving settings, the term `hammer' is often used to indicate built-in automation (of various kinds, depending on the interactive proving system used). There has been recent work on learning for which proof states to invoke other automation in the Thor project~\citep{DBLP:conf/nips/JiangLTCOMWJ22}.

Aside from the machine learning guidance on the ITP level, there has also been work on improving the performance of first-order automated theorem provers (and SMT solvers), that form the foundation for the `hammer' approaches, on ITP-derived problems. Work on the `hammer' approach in various settings was surveyed in \citet{DBLP:journals/jfrea/BlanchetteKPU16}. The ENIGMA system~\citep{enigmaoriginal} was originally used on first-order translated versions of problems from the Mizar system, there were several improvements on the Mizar version \citep{enigmang,enigmaanon}, including a graph-neural-network-based version. A similar system was also trained and tested on Isabelle problems \citep{DBLP:conf/itp/GoertzelJKOPU22}. 

%\stepcounter{section}
% \section{Example interaction session}

% \subsection{Example of proof with new definitions}

% \subsection{Example of shorter / better proofs}

\section{The Universe of Mathematical Concepts}\label{sec:big-graph}

The dataset we utilize is extracted from 120 different Coq packages from the
Opam package manager. These packages were selected by a SAT solver as the
largest mutually consistent set of packages available for Coq v8.11. Their
topics vary wildly, including analysis, compiler and programming language
formalization, separation logic, homotopy type theory, and much more. The graph
extracted from these formalizations consists of over 250 million nodes, which
encode 520k definitions, of which 266k are theorems, and 4.6M proof state
transformations\footnote{Roughly half of the definitions are derived from each
  other through Coq's module and section mechanism.}.

We divide packages into training and test where no test package depends on a
training package. To do so, we induced a random topological order on the Coq
packages, with regard to the dependency graph. The resulting list was then split
such that the average percentage of theorems and of proof states in the training
split is close to 90\% (in our case, it is 91.3\% of all theorems and 88.8\% of
all proof states).

\begin{figure*}[!htbp]
  \centering
  \includegraphics[width=\textwidth]{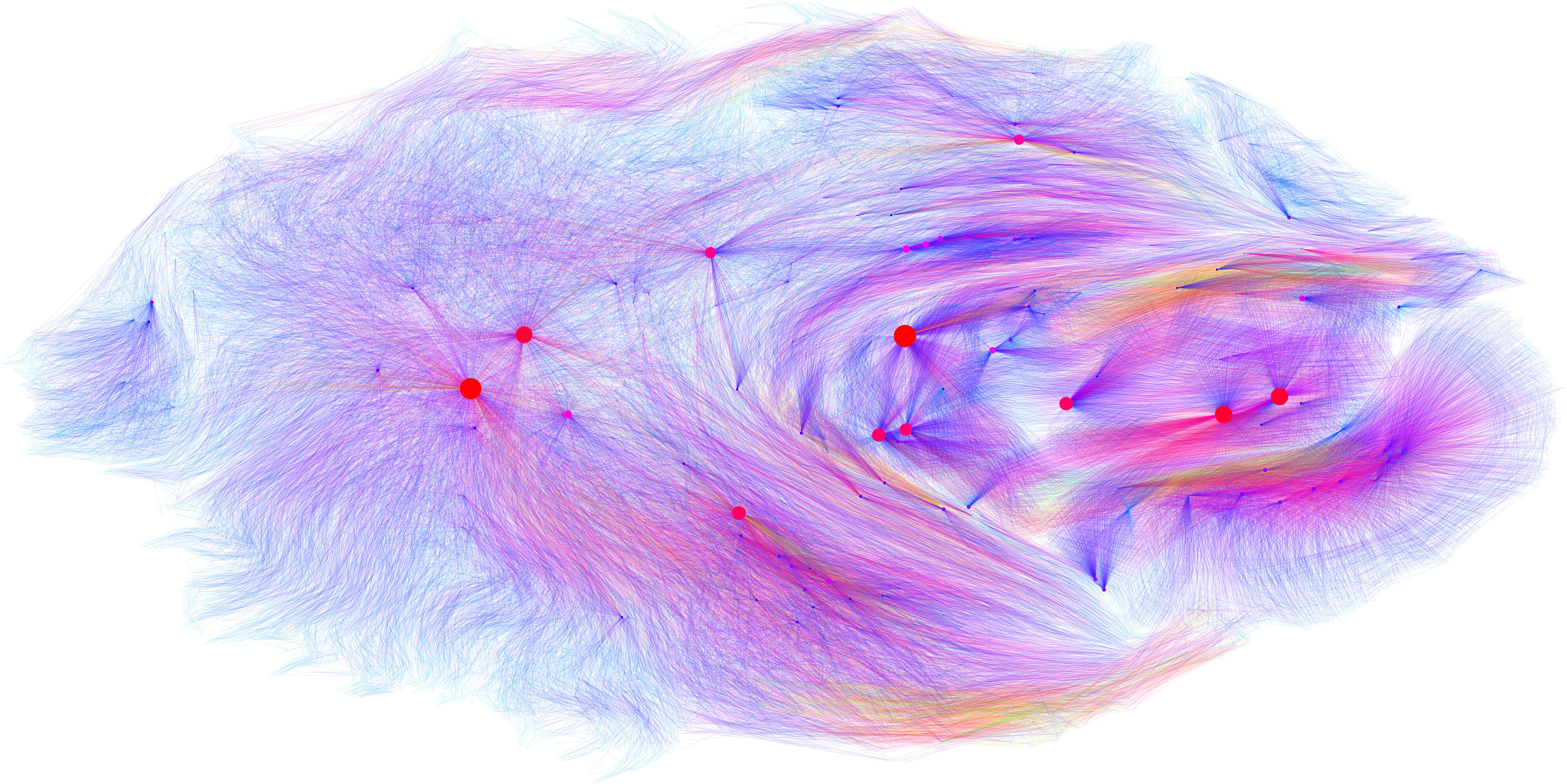}
  \caption{A rendering of a small section of the known mathematical universe as
    formalized in the Coq proof assistant.}
  \label{fig:web}
\end{figure*}

Figure~\ref{fig:web} shows a rendering of a small section of the dataset. In
particular, only the most basic of mathematical concepts that are part of the
``Prelude'' in Coq's standard library, are rendered. The full graph contained in
the dataset would be over 3000 times larger. The graph includes all definitions,
lemmas, and proof terms in the Prelude, as well as a representation of all
intermediate proof states that were generated during the proving process and the
partial proof terms generated by tactics to transition between proof states.
Together, they form a single, interconnected graph. The graph is fully
de-duplicated, such that every sub-term is represented only once (but may be
referenced many times).

The size and color of a node in the rendering is dependent on how many times it
is referenced. As nodes get more popular, they increase in size. Examples of
large nodes are the universes Prop and Type, the inductive definition for
Leibnitz equality and the inductive definitions for true and false. Not all
popular nodes are definitions, however. Other popular mathematical entities
include hypotheses that posit the existence of natural numbers and booleans, and
even some anonymous subterms that happen to occur very often.

The placement of each node is determined by the \textit{sfdp} force-directed graph drawing
engine of the Graphviz visualization suite~\citep{hu2005efficient}. As a result,
related nodes will be rendered close together. This is particularly apparent for
inductive definitions and their constructors, where it often appears as if the
constructors are ``planets'' that orbit around a corresponding inductive
that acts as their ``sun''. The color of each edge is determined by its length
as determined by the rendering engine.

\section{Further Details on the G2T Architecture}
Here we give further details about the G2T architecture described in Section~\ref{sec:gnn-def-task}.

\paragraph{Graph Inputs for Definitions and Proof States}
The inputs to the definition and prediction task are directed graphs.
In Figure~\ref{fig:neural_network}, the input to the definition task
is a graph corresponding to the definition $f\,x := x+x$.
The graph of $f$ can be seen in Figure~\ref{fig:graph-example}
as the eight node subgraph corresponds to all the green edges
and all the nodes touching the green edges.
This includes the root node $f$,
the global definition nodes for dependencies $\mathbb{N}$ and $+$,
and a number of other non-definition nodes
for binders ($\forall$ and $\lambda$),
for function application ($@$),
and for the variable $x$ shown as $\uparrow$.
The edge labels come with labels that are not shown.
Variables do not need names since they are connected to their binders via labeled edges.
Even though we include the nodes
$\mathbb{N}$ and $+$, 
we do not include the nodes in their respective definition graphs.
This keeps the size of our input graph manageably small.

Similarly, in Figure~\ref{fig:neural_network},
the proof state input to the prediction network corresponds to a proof state
with two local hypotheses $x: \mathbb{N}, y: \mathbb{N}$
and the goal $\vdash 2*x=f\, x$.
This is almost the subgraph shown with blue edges in Figure~\ref{fig:graph-example}
except it contains an extra local hypothesis $y$.
Like variables, we do not store the names of local hypotheses in the graph.
The triangle $2*x$ represents the subgraph of nodes shared between the proof state and $T_1$.
(To compress our graph, common terms are shared.)
The nodes in that triangle are also part of the proof state graph.

In step A of Figure~\ref{fig:neural_network}, we assign embeddings as follows:
\begin{itemize}
\item Root nodes like $f$ get a null vector of all zeros.
\item Definition nodes like $2$, $*$, $+$, and $\mathbb{N}$
are assigned embeddings from the definition embedding table.
\item Other nodes, are assigned embeddings from the node embedding table
based only on the node type.
These include nodes for binders, function application, variables, and local constants.
\item Edges are assigned embeddings according to their edge types (after adding in extra edges as described in Section~\ref{sec:gnn-def-task}).
\end{itemize}

While Step A treats local context nodes the same as any other non-definition node,
they become important in Step G,
where we use the GNN embeddings for the local context nodes to select arguments.

\paragraph{RNN for Local and Global Argument Prediction}
The RNN described in Step F of Figure~\ref{fig:neural_network}
is a simple two layer RNN described mathematically as follows.
Let $t$ be the embedding of the base tactic (\verb|apply _| in Figure~\ref{fig:neural_network}).
Let $p$ be the pooled embedding of the proof state,
that is the mean of all the node embeddings in the output graph of the GNN.
The goal is to compute an embedding $x_i$ for each argument position $i$
in the base tactic.
(For example, \verb|rewrite _ in _| has two argument positions,
\verb|reflexivity| has zero positions, and
\verb|apply _| has one position.
Hence in the example in Figure~\ref{fig:neural_network}, there is only one RNN output.)
The computation is as follows:
\begin{align*}
x^{(0)}_i &:= p &\\
h^{(n)}_i &:= t\qquad&\text{for }n \in \{0, 1\} \\
x^{(n+1)}_i, h^{(n)}_{i+1} &:= \text{ReLU}(\text{Dense}(x^{(n)}_i, h^{(n)}_i))
\end{align*}
The two inputs and two outputs to the dense layer are concatenated.
The RNN output corresponding to the $i$th argument position is $x^{(2)}_i$.
This is used as the input to Step G in Figure~\ref{fig:neural_network}.

\paragraph{The Loss for Definition and Prediction task}
Given a definition $d$ with index $i_d$ and graph $g_d$,
let $\text{DefTask}(g_d)$ be the computed embedding for the definition task,
and $\text{DefEmb}(i_d)$ be the embedding in the definition embedding table.
The definition loss is the cosine similarity loss
\begin{align*}
\mathcal{L}_{\text{def}} &:= 1-\frac{\text{DefTask}(g_d) \cdot \text{DefEmb}(i_d)}{\left|\text{DefTask}(g_d)\right|\,\left|\text{DefEmb}(i_d)\right|}\\
&= 1-\text{DefTask}(g_d) \cdot \text{DefEmb}(i_d)
\end{align*}
where $\cdot$ is inner product.  The denominator is not needed since both $\text{DefTask}(g_d)$ and $\text{DefEmb}(i_d)$ are unit normalized.

The prediction task loss is calculated as follows.
Let $P_{\text{tactic}}(T\mid S)$ be the probability the model assigns to a base tactic $T$
given a proof state $S$.
Let $P_{\text{arg}_i}(A\mid S, T)$ be the probability the model assigns to an argument $A$
(either local or global) given a proof state $S$ and the base tactic $T$.
(Notice, even though we use an RNN, the probability of the $i$th argument only depends on the
tactic and proof state.)
Then if the ground truth tactic sequence is the base tactic $T$ with arguments $A_0, A_1, ..., A_{n-1}$, the loss is calculated as the cross entropy
\begin{align*}
\mathcal{L}_{\text{tactic}} := &-\log P_{\text{tactic}}(T\mid S)
-\log P_{\text{arg}_0}(A_0\mid S, T)\\
                               &-\log P_{\text{arg}_1}(A_1\mid S, T) - ... \\
                               &-\log P_{\text{arg}_n}(A_{n-1}\mid S, T)
\end{align*}

\paragraph{Adding New Definitions to the Current State}
During inference, every time we get to a new theorem,
we query Coq (via Tactician) for any new definitions added to the global context.
This gives us a sequence of definitions $d_0, ..., d_{n-1}$.
For each definition $d$,
we first check if that definition was seen during training
(using a hash based on the graph representation).
If so, we align the definition index given by Coq to the one we already have.
If not, we use $\text{DefTask}(g_d)$ to calculate the definition embedding,
where $g_{d}$ is the graph of definition $d$
and $\text{DefTask}$ is our definition task neural network.
All the embedding tables are in the neural network,
including the definition embedding table.
We may assume, by the order definitions are returned from Coq,
that any prerequisite definitions 
needed to calculated $\text{DefTask}(g_d)$
have already been added to the definition embedding table.
After calculating $\text{DefTask}(g_d)$,
we manually update the definition embedding table
to include the newly calculated embedding for $d$.

As an implementation detail, the most expensive part of changing the embedding table
is recompiling the network, and this is only needed when we change the size of the table.
To reduce this during inference, we allocate a larger definition embedding table than needed,
similar to a resizable array.

\section{Training Details}
Graph2Tac graph models were trained on 8 CPUs (with 128 GB RAM) and 2 NVIDIA A100 GPU cards for about 20 days.
Optimization used the Adam optimizer with a learning rate of 0.0003
and L2 regularization set to 1.0e-05.
Training was restarted every 24 hours from the current checkpoint,
but we did not save optimizer states between restarts.
We noticed periodic instability in training and occasionally
had to restart training from a checkpoint.

During the design of the model, some hyperparameters, such as the coefficient $1000$ for the loss,
were tuned manually on a small subset of the training data using other
training packages for validation.

The transformer was a GPT2-style model~\citep{radford2019language-gpt2} from the Huggingface Transformers library. The model was trained to predict the tactics based on the proof states. To determine whether the model was sufficiently trained, we monitored the validation accuracy and stopped training when the validation performance stagnated.

The GPU evaluations and the training of the transformer were performed on machines with two 20-core Inter
E5-2698v4, a working memory of 512 GB RAM, and 8 NVIDIA Tesla V100 GPU cards.

\section{Search Procedure Details}\label{sec:search-procedure}
The G2T, transformer, and $k$-NN \emph{models} only predict single tactics.
In order to turn our models into a \emph{solver},
which can find a whole proof,
we combine them with the search procedure of Tactician~\citep{DBLP:conf/lpar/BlaauwbroekUG20},
namely Dijkstra's algorithm%
\footnote{Dijkstra's algorithm is a special case of
best-first search algorithms,
and the "best-first search" of \citet{polu2020generative}
and following works \citep{han2021proof, DBLP:conf/nips/JiangLTCOMWJ22}
is equivalent to vanilla Dijkstra's algorithm
when the search nodes are ordered by
the cumulative log probability of the tactic predictions.} %
to explore the tree of possible proofs, 
but modified to use iterative deepening instead of a queue.
For G2T and transformer, each possible tactic is scored by
the log probability predicted by the corresponding model.
The length of a path (partial proof) is the
sum of the negative log probabilities of its steps.
With iterative deepening,
each iteration $i$ only explores paths
whose distance is below an upper bound $D_{\text{max}}^i$ (depth first where the tactics are ordered by score).
If an iteration exhausts all valid paths without finding a proof,
the search starts from the beginning with the new upper bound
\[
    D_{\text{max}}^{i+1} = D_{\text{max}}^{i+1} + 1 + D_{extra}^i
\]
where $D_{extra}^i$ is the negative log probability
needed to increase the left-most path by one tactic.
Our iterative deepening search uses constant memory
at the expense of only a constant multiplicative increase in time.

A Coq tactic can produce multiple independent subgoals.
Each step in our search procedure predicts a tactic
for the first goal of the stack
and applies the resulting tactic to that goal only.
If these goals are fully independent (have no shared E-variables),
then our DFS takes this into account by not backtracking
into a previously solved subgoal earlier in the proof.
See Tactician~\citep{DBLP:conf/lpar/BlaauwbroekUG20} for more details.\todo{I think I understand this now, and if needed I could give an example in the appendix. -Jason}

\section{All Results}
Table~\ref{tab:10min_table} shows the results for the 10-minute evaluation
where packages \emph{hott} and \emph{tlc} are excluded.

\begin{figure*}[!htbp]
    \centering
    \includegraphics[width=0.9\textwidth]{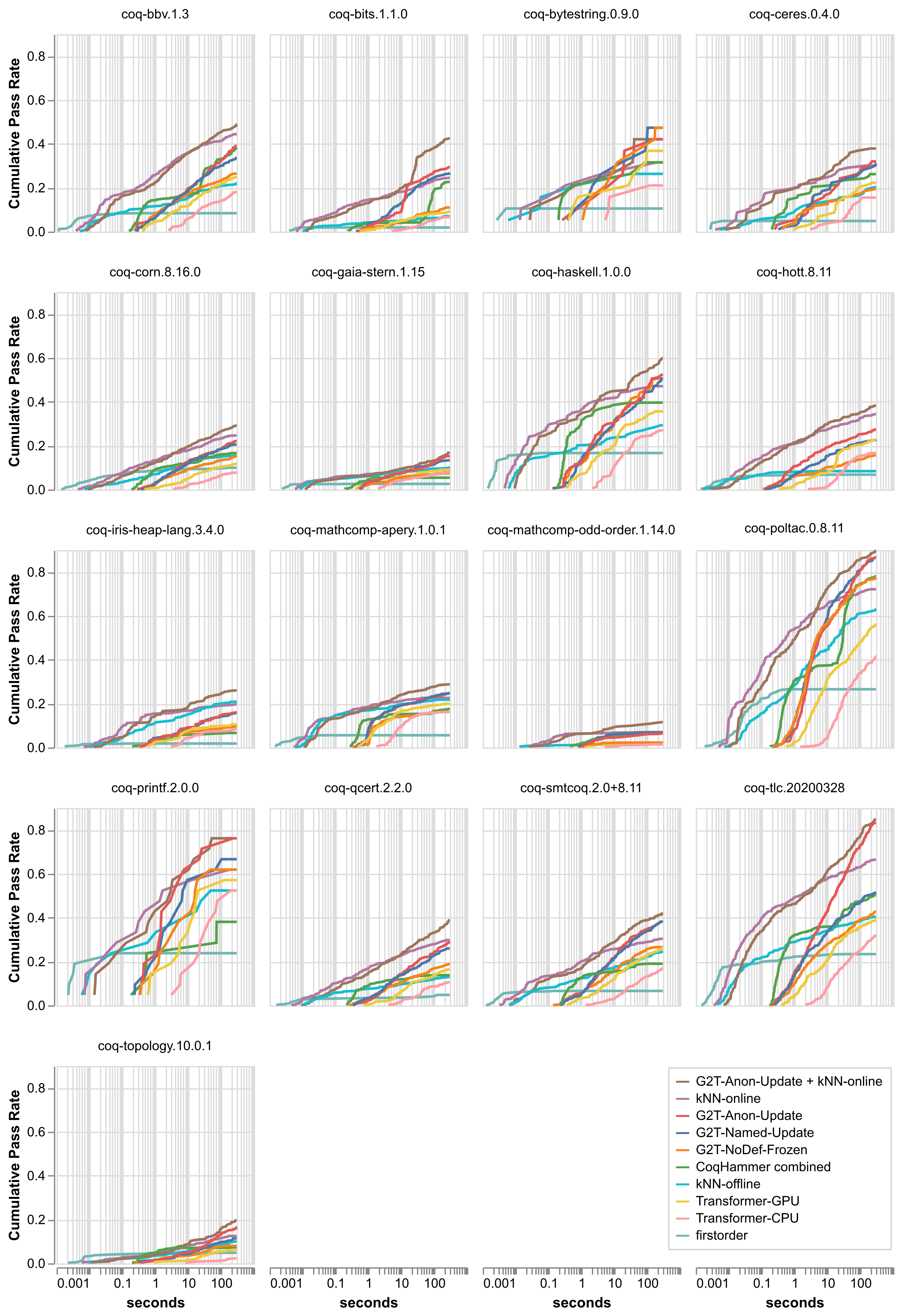}
    \caption{Cumulative pass rates for all packages.}
    \label{fig:grid-plot-all-packages}
\end{figure*}

Figure~\ref{fig:grid-plot-all-packages} shows the cumulative pass rates for all packages and all the models discussed in Section~\ref{sec:evaluation}.
We see that neither the G2T-Anon-Update nor the online $k$-NN always performs
better.
In many cases G2T-Anon-Update either overtakes the online $k$-NN or appears as if it will overtake the online $k$-NN if given enough time.
As discussed in Appendix~\ref{sec:axioms},
the \emph{tlc} package is special in
that G2T-Anon-Update was able to prove a significant number of
theorems using an inconsistent axiom.

\begin{table}[!htbp]
\centering
\begin{tabular}{lr}
\toprule
solver & pass rate \\
\midrule
CoqHammer combined & 0.174 \\
G2T-Anon-Update & 0.261 \\
G2T-Anon-Update + $k$NN-online & 0.332 \\
G2T-Named-Update & 0.241 \\
G2T-NoDef-Frozen & 0.175 \\
Transformer-CPU & 0.105 \\
Transformer-GPU & 0.148 \\
firstorder & 0.075 \\
$k$NN-offline & 0.150 \\
$k$NN-online & 0.258 \\
\bottomrule
\end{tabular}
\caption{Pass rates for 10-minute time limit.}
\label{tab:10min_table}
\end{table}

\newpage

\section{Informal Comparison on CoqGym Test Packages}\label{sec:CoqGym}

Besides highlighting our novel definition task,
our work is the first work we are aware of comparing
$k$-NN, transformer, and graph-based solvers for an
interactive theorem prover.
Perhaps surprisingly, the online $k$-NN performs much better
than expected, especially since most recent papers use powerful 
neural models like transformers.

The primary benchmark for Coq is the CoqGym benchmark,
consisting of 28 test projects.
Unfortunately, of those 28, only seven are included in our dataset
(our data is harvested from Opam packages, while the CoqGym data is
retrieved from Github).
Of those seven projects, all but one (\textit{poltac}) are included in our training set,
making it impossible to do a fair evaluation with any of our trained models.
Nonetheless, we can compare our trained models on \emph{poltac} and
our untrained models, such as the online $k$-NN model, on five of the remainder.\footnote{
We excluded \emph{goedel} as it has completely changed between the CoqGym
benchmarks and ours.
It has been abandoned and merged into \emph{hydra-battles}.
Our package of \emph{goedel} only contains 17\% of the 606 theorems found
in the older \emph{goedel} package in CoqGym.}

\begin{table*}[!htbp]
\centering
\footnotesize
\setlength\tabcolsep{4pt}

\begin{tabular}{lr|ccccc|c}
\toprule
& time & \multicolumn{6}{c}{pass rate (solved/total)} \\
& (min) & buchberger & coquelicot & hoare-tut & huffman & zorns-lemma & total\\
\midrule
ASTactic & 10 & 0.097 \hfill (70/725) & 0.065 \hfill (95/1467) & 0.056 \hfill (1/18) & 0.080 \hfill (25/314) & 0.067 \hfill (10/149) & 0.105 \hfill (319/3036) \\
CoqHammer & 10 & 0.229 \hfill (166/725) & 0.186 \hfill (273/1467) & \textbf{0.333} \hfill (6/18) & \textbf{0.236} \hfill (74/314) & 0.121 \hfill (18/149) & 0.272 \hfill (826/3036) \\
Diva & 64$\times$10 & 0.204 \hfill (148/725) & 0.128 \hfill (188/1467) & 0.278 \hfill (5/18) & 0.178 \hfill (56/314) & 0.121 \hfill (18/149) & 0.193 \hfill (585/3036) \\
TacTok & 2$\times$10 & 0.105 \hfill (76/725) & 0.068 \hfill (100/1467) & 0.278 \hfill (5/18) & 0.089 \hfill (28/314) & 0.081 \hfill (12/149) & 0.110 \hfill (333/3036) \\
TacTok+Passport & 2$\times$10 & 0.126 \hfill (91/725) & 0.082 \hfill (120/1467) & 0.222 \hfill (4/18) & 0.096 \hfill (30/314) & 0.081 \hfill (12/149) & 0.129 \hfill (392/3036) \\
\midrule
ProverBot9001 & NA & 0.237 \hfill (178/750) & 0.107 \hfill (187/1743) & 0.280 \hfill (7/25) & 0.225 \hfill (71/316) & 0.115 \hfill (18/156) & 0.194 \hfill (639/3299) \\
\midrule
CH combined & 10 & \textbf{0.284} \hfill (213/750) & 0.148 \hfill (258/1743) & 0.320 \hfill (8/25) & 0.207 \hfill (65/314) & 0.100 \hfill (26/259) & 0.242 \hfill (822/3400) \\
$k$NN-online & 10 & 0.264 \hfill (198/750) & \textbf{0.234} \hfill (408/1743) & 0.320 \hfill (8/25) & 0.169 \hfill (53/314) & \textbf{0.282} \hfill (73/259) & \textbf{0.284} \hfill (967/3400) \\
\bottomrule
\end{tabular}

\caption{Comparison of results across three families of benchmarks.
Note each family has a different number of theorems.
The total column includes the \emph{poltac} results in Table~\ref{tab:coqgym_comparison2}.
The Diva and TacTok solvers run multiple solvers in parallel, each for 10 minutes.
}
\label{tab:coqgym_comparison1}
\end{table*}

\begin{table}[!htbp]
\centering
\footnotesize
\setlength\tabcolsep{4pt}

\begin{tabular}{lr|c}
\toprule
& time & \multicolumn{1}{c}{pass rate (solved/total)} \\
& (min) & poltac \\
\midrule
ASTactic & 10 & 0.325 \hfill (118/363) \\
CoqHammer & 10 & 0.796 \hfill (289/363) \\
Diva & 64$\times$10 & 0.468 \hfill (170/363) \\
TacTok & 2$\times$10 & 0.309 \hfill (112/363) \\
TacTok+Passport & 2$\times$10 & 0.372 \hfill (135/363) \\
\midrule
ProverBot9001 & NA & 0.576 \hfill (178/309) \\
\midrule
CoqHammer combined & 10 & 0.816 \hfill (252/309) \\
G2T-Anon-Update & 5 & 0.864 \hfill (267/309) \\
G2T-Named-Update & 5 & \textbf{0.867} \hfill (268/309) \\
G2T-NoDef-Frozen & 5 & 0.773 \hfill (239/309) \\
Transformer (GPU) & 5 & 0.560 \hfill (173/309) \\
$k$NN-online & 10 & 0.735 \hfill (227/309) \\
\bottomrule
\end{tabular}

\caption{Comparison of results on \emph{poltac},
which is in both our test set and the CoqGym test set,
across three families of benchmarks.}
\label{tab:coqgym_comparison2}
\end{table}

In Tables~\ref{tab:coqgym_comparison1} and \ref{tab:coqgym_comparison2},
we compare our work to two other families of benchmarks on the CoqGym test projects.
The first family are benchmarks of \citet{yang2019coqgym}, \citet{tactok}, \citet{diva}, and \citet{passport}.
These include the neural solver ASTactic and its extensions TacTok, Diva, TacTok+Passport,
as well as an independent benchmark of CoqHammer.
The second is a benchmark of the neural solver Proverbot9001 \citep{proverbot9001}.
While the original ProverBot9001 results were for CompCert,
we use data from the ProverBot9001 benchmark found at \url{https://proverbot9001.ucsd.edu/compare/}.
It is a comparison to CoqGym using approximately similar training and test packages, but run in the ProverBot9001 system.
For our family of benchmarks, we include both
the trained solvers reported in our results section (only tested on \emph{poltac}),
as well as a full 10 minute online $k$-NN and CoqHammer benchmark run on all theorems in all six projects.

It should be emphasized that these comparisons are informal
and non-conclusive due to a number of methodological differences:
\begin{itemize}
\item The ASTactic/TacTok/Diva and ProverBot9001 benchmarks train on the CoqGym training suite, or in the case of ProverBot9001 an approximation of the CoqGym training suite, while our G2T models train on our training suite containing different Coq projects.
\item Coq versions used by the benchmarks vary between version 8.10 and 8.12.
\item Each benchmark may use different versions of test packages,
which includes different (in some cases significantly different) number of test theorems.
\item The ASTactic/TacTok/Diva family of benchmarks were run for 10 minutes,
although TacTok and Diva run multiple solvers in parallel
for a total run time, respectively, of 2$\times$10 and 64$\times$10 minutes.
Some of our benchmarks were run for 5 minutes and some for 10.
ProverBot9001, on the other hand, does not use a time limit,
instead stopping the search based on search depth.
Nonetheless, the ProverBot9001 results were specifically designed by the creators
to compare to the ASTactic/TacTok family of benchmarks.
\item There are differences in hardware.
ASTactic, for example, uses two cores, where we restrict ourselves to one core.
Since ProverBot9001 does not use time limits, hardware does not matter.
\end{itemize}

Nonetheless, the previous CoqGym benchmarks have a similar flavor to ours.
All results test on entire packages not seen during training.
The benchmarks were all run without a GPU (except our transformer baseline).
None of the systems use large-scale pre-training on internet data,
so there is little risk of data leakage.

The results in Table~\ref{tab:coqgym_comparison1}
and Table~\ref{tab:coqgym_comparison2} suggest that the online $k$-NN model,
despite being one of the simplest models we compare, is one of the strongest.
In particular, it outperforms the other previously published neural models,
only doing worse than ProverBot9001 on one project. 

Transitively, this suggests our best G2T models are also state of the art for Coq,
and we indeed see this on \emph{poltac} in Table~\ref{tab:coqgym_comparison2}.

Finally, note that our CoqHammer combined aggregate solver
performs similarly to the previous benchmark of CoqHammer,
suggesting our aggregate is a good approximation (see Section~\ref{section:CoqHammer}).
Nonetheless, our CoqHammer benchmarks and those in CoqGym likely
use different versions of CoqHammer and
likely rely on different combinations of ATP backends.

In conclusion, more work is needed to compare the methods in this paper to other methods in the literature.
The strong showing of the online $k$-NN baseline,
which until now has not been properly compared to other systems,
suggests that many of the approaches in the literature,
especially the common transformer approaches,
may be surprisingly weak,
especially in the online setting of new files or packages not seen during training.

\section{Analysis of CoqHammer Benchmarks}\label{section:CoqHammer}
A core design principle of CoqHammer is the ability to translate a higher-order
theory into a first-order theory and export it to a varied set of automatic
theorem provers. Data provided by the original CoqHammer
publication~\citep{coqhammer} suggests that the various ATP's are highly
complementary. Indeed, this is confirmed in a publication on
Tactician~\citep{DBLP:conf/lpar/BlaauwbroekUG20} which compares an early version
of its online $k$-NN solver with CoqHammer on Coq's standard library. The online $k$-NN solver
is able to beat any of CoqHammer's individual ATP backends by some margin. But
when combined, the ATP's achieve a higher pass-rate than the online $k$-NN solver.

\begin{figure}[!htbp]
    \centering
    \includegraphics[width=1.0\columnwidth]{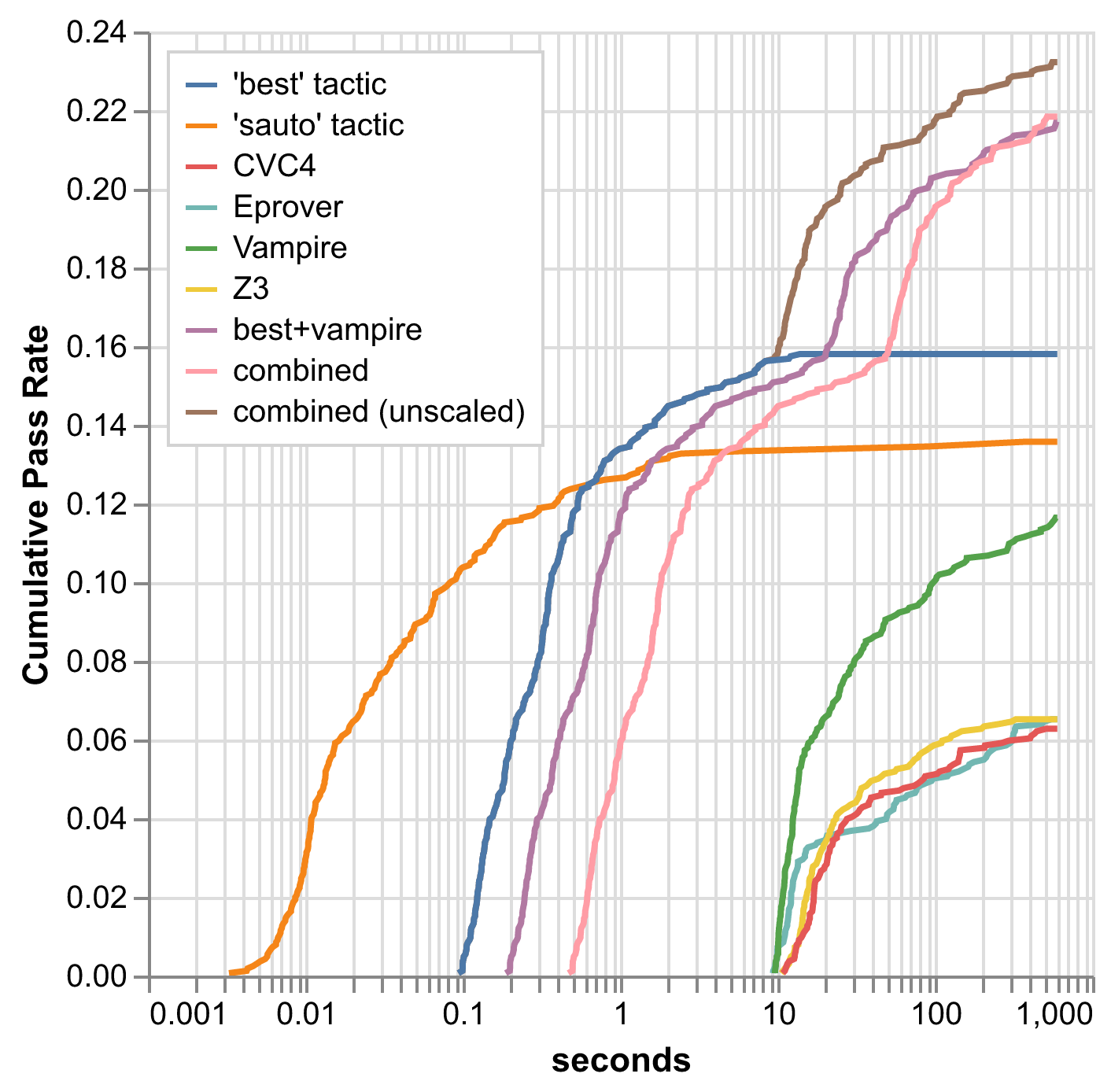}
    \caption{Pass rates for individual solvers of CoqHammer and various aggregates.
    (Includes all test packages except \emph{hott}.)}
    \label{fig:coqhammer-times}
\end{figure}

\begin{figure}[!htbp]
    \centering
    \includegraphics[width=1.0\columnwidth]{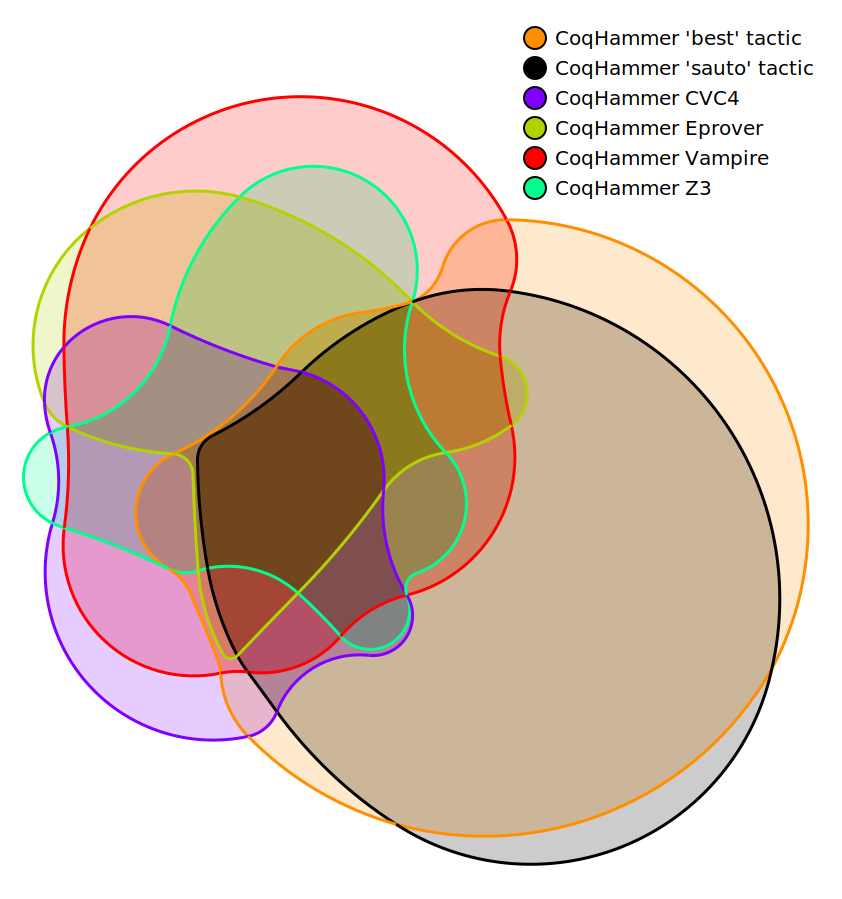}
    \caption{Venn diagram showing complementarity between CoqHammer
      solvers.
      (Includes all test packages except \emph{hott}.)}
    \label{fig:coqhammer-venn-diagram}
\end{figure}

To obtain a fair comparison, we benchmarked all of CoqHammer's ATP backends
individually. We then calculate a time-scaled combination of the ATP's in order
to simulate the parallel execution of all ATP's on a single CPU. Surprisingly,
our results are quite different from what is reported in previous
publications.\footnote{Note that we are unable to reproduce the
  benchmark results on Coq's standard library because CoqHammer depends on the
  same standard library. This creates a circularity that cannot easily be broken.}
As can be seen from the Venn diagram in Figure~\ref{fig:coqhammer-venn-diagram},
the ATP's are not nearly as complimentary on our testing set as one would
expect. The Vampire backend is able to prove most of the theorems obtained by
other ATP's. Only the \texttt{sauto} and \texttt{best} tactics remain highly
complementary to Vampire. Figure~\ref{fig:coqhammer-times} shows that a
combination of \texttt{best} and Vampire produces the same results as an
aggregation of all solvers, but faster. As such, unless one truly has
CPU-threads to spare for parallel solving, using many ATP's is not effective.

As a result, CoqHammer is no longer able to beat the online $k$-NN and G2T solver on our
testing set. We speculate that the root cause of this discrepancy is the use of
a more complex test set, on which the higher-order to first-order translation of
CoqHammer is far less effective. Presumably, during the development of
CoqHammer, the translation was tuned particularly well on Coq's standard
library. This translation appears to generalize to other packages less well than
one would hope.


\newpage

\section{The Relative Speed of the Models}
\begin{figure}[!htbp]
    \centering
    \includegraphics[width=1.0\columnwidth]{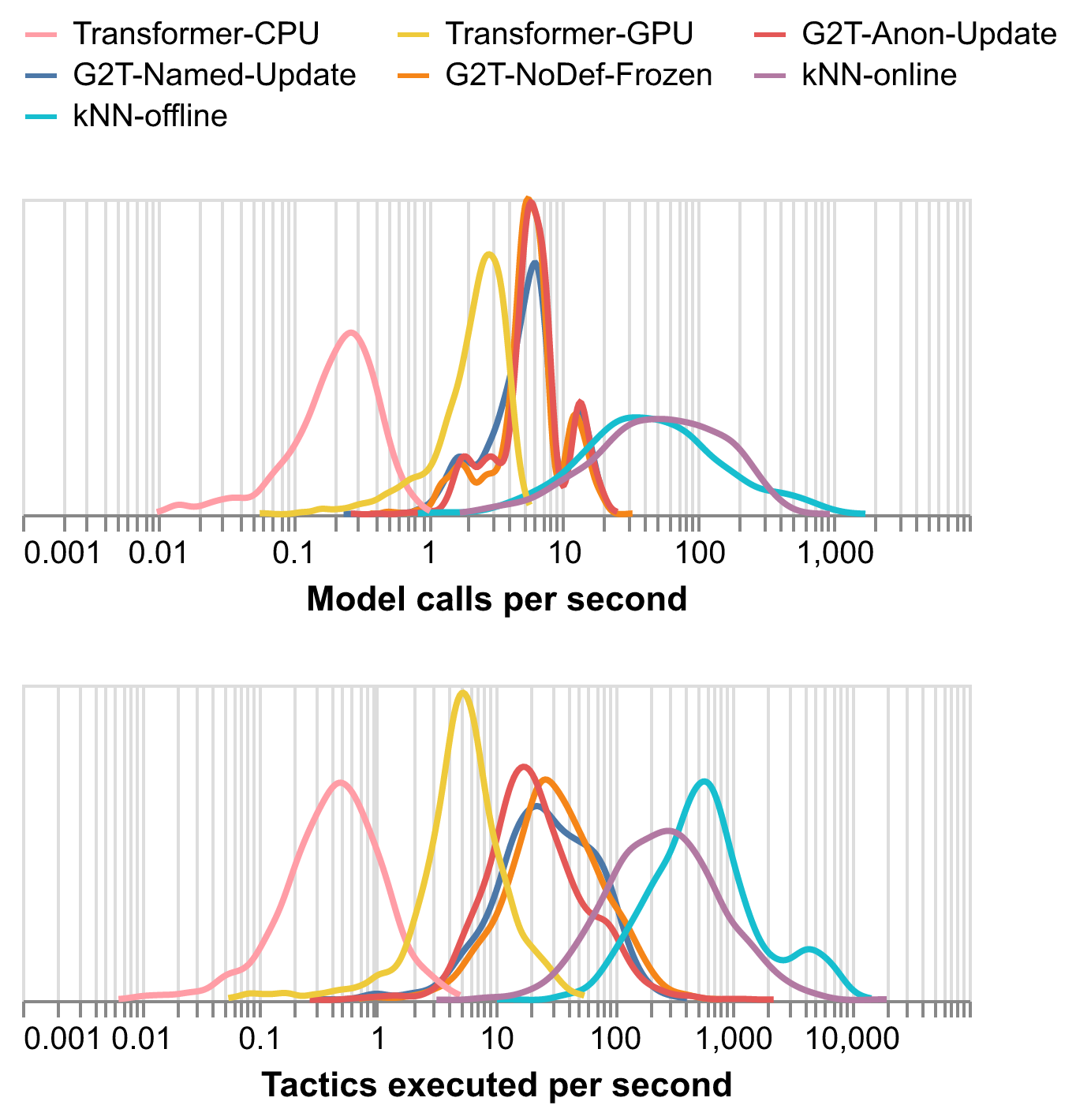}
    \caption{Density plots of the speed of the various models, in tactics executed per second and model calls per second.}
    \label{fig:model-speeds}
\end{figure}
Figure~\ref{fig:model-speeds} shows the speed of the models for the attempted proofs in the test set.  Notice the various families of models can differ by an order of magnitude in speed, with $k$-NN models being the fastest, and transformers being the slowest.

\begin{table*}[ht!]
\centering
\begin{tabular}{lllll}
\toprule
Solver & \multicolumn{2}{c}{Global context proof data} & LSHF & Notes \\
 &  files & recent & & \\
\midrule
$k$NN-recent-online       & imports + current file   & 1000 &            & ``online $k$-NN'' in results \\
$k$NN-LSHF-online          & imports + current file   & all  & \checkmark & $k$-NN in Tactician \\
$k$NN-recent-allButFile  & imports                  & 1000 &            & \\
$k$NN-LSHF-allButFile     & imports                  & all  & \checkmark & \\
$k$NN-recent-offline & imports in training data & 1000 &            & \\
$k$NN-LSHF-offline    & imports in training data & all  & \checkmark & ``offline $k$-NN'' in results \\
\bottomrule
\end{tabular}
\caption{$k$-NN models in the ablation.}
\label{tab:knn_table}
\end{table*}

\section{Ablation of $k$-NN Solvers Seeing Recent Proofs}\label{sec:kNN-ablation}

In this section, we explore variations of the $k$-NN setup.
The online and offline $k$-NN solvers in our main results
have multiple differences from each
that determine which examples in the global context
the $k$-NN can search over.
The online $k$-NN can search over only the most recent 1000 tactics,
but can see proofs up to and including those in the current file.
The offline $k$-NN is not restricted to 1000 tactics,
but can see all the tactics that were available at training time,
and uses locality sensitive hashing forests (LSHF)
to account for the larger search space.

To break down the contribution of each difference,
in Table~\ref{sec:kNN-ablation},
we specify six solvers that vary on two dimensions.
The first dimension is how online the model is.
The most online models search over proofs up to and including the current file.
The next level excludes the current file.
The final level excludes all but the training data.

The second dimension is whether the model only sees the 1000 most recent tactics.
For proofs in different files, the order of the theorems is someone arbitrary,
and this likely affects the results.
If we are not restricting to 1000 tactics,
then we use LSHF to speed up the search.

The solver $k$NN-LSHF-online is the default $k$-NN solver in Tactician
\citep{DBLP:conf/mkm/BlaauwbroekUG20}.
The solvers $k$NN-recent-online and $k$NN-LSHF-online
are the solvers in the results section.

\begin{figure}[!htbp]
    \centering
    \includegraphics[width=1.0\columnwidth]{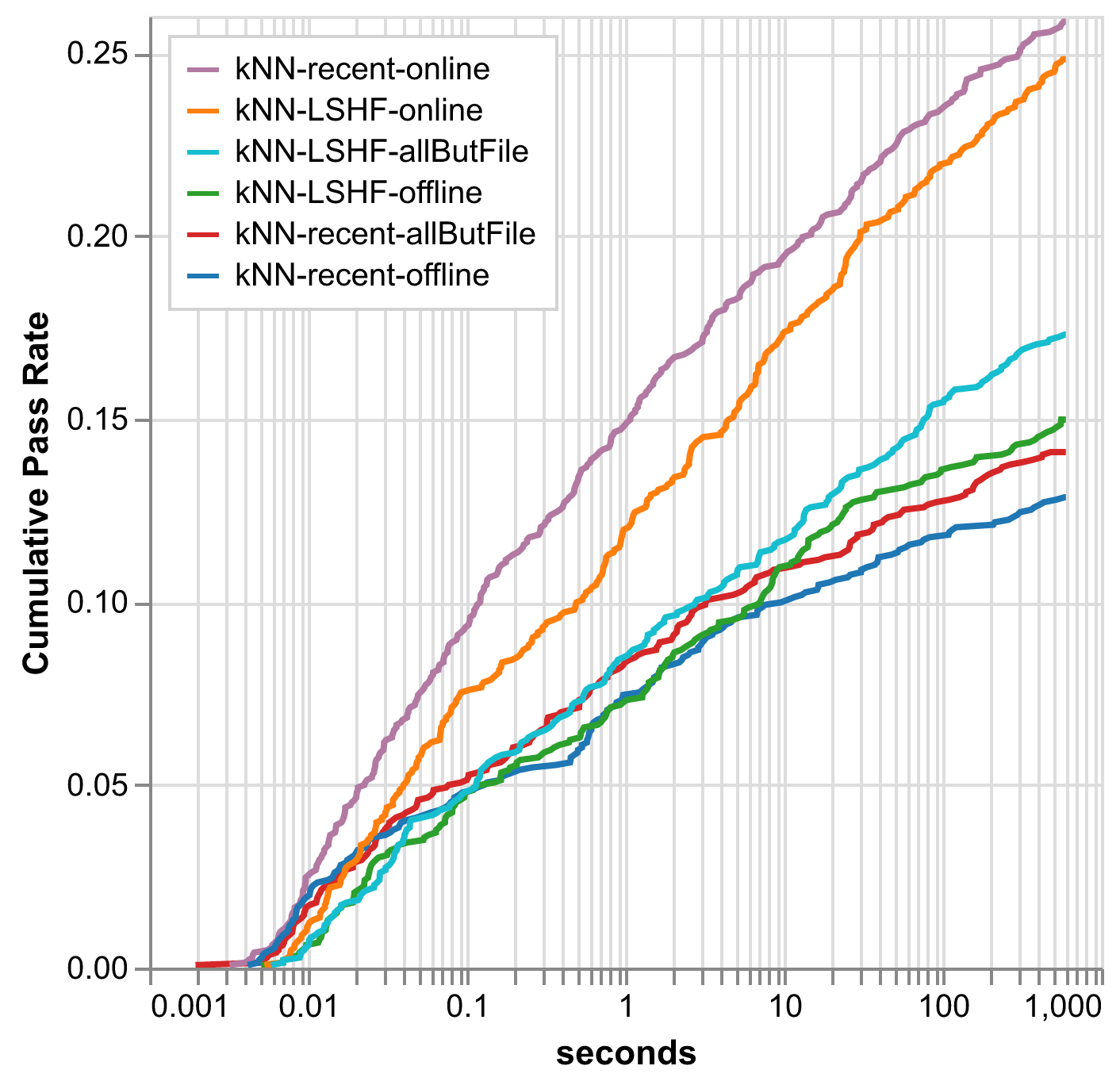}
    \caption{Pass rates for the different $k$-NN solver variations.}
    \label{fig:knn-test-packages-time}
\end{figure}

Figure~\ref{fig:knn-test-packages-time}
gives the performance of all six solvers.
There is a significant performance advantage for the full online solvers,
which can see previous proofs in the same file,
and even a bit more advantage when the model is restricted to
\emph{only} the most recent 1000 tactics.
This suggests that our online $k$-NN ($k$NN-recent-online),
gets much of its power from copying nearby tactics of proofs in the same file.

There is also some smaller advantage
when going from a model that can see only the training data
to one that can see all but the most recent file.
This again shows the importance of recent online information.

The worst models are the ones restricted to the most "recent" 1000 tactics,
but where they cannot see the current file.
This is likely an artifact of the arbitrary
way that proofs are ordered outside the current file.

Overall these results suggest the online $k$-NN approach
may be slightly brittle in that it depends a lot on what recent
proof data is available in the global context.
Nonetheless, given the speed and simplicity of the $k$-NN solvers,
and how even the offline $k$-NN models have non-negligible performance,
we think they are good baseline models for theorem proving.

\section{Effect of Training Time on G2T Solvers}\label{sec:training-time}
The G2T-Anon model was trained for 247 epochs over 20 days.
In this section we show that we can achieve comparable results
with only 27 epochs in 2 days.
In table~\ref{tab:training-time} and Figure~\ref{fig:gt2-test-packages-time},
we see that a model trained for only 27 epochs
(the earliest checkpoint we saved)
does as well, if not better, than a model trained for the full 247 epochs,
despite the fact that the latter model has better validation accuracy.
(Moreover, we found this behavior was consistent across packages.)
This is important for the following reasons:
(1) Faster model training allows for more rapid development and experimentation.
(2) During training, after about 100 epochs we regularly encounter instabilities
requiring us to restart training from a previous checkpoint.
(3) Shorter training time likely means shorter fine-tuning,
hopefully making it possible for end users to fine-tune our model on new Coq projects,
even with limited GPU availability.

\begin{table}[!htbp]
\centering
\addtolength{\tabcolsep}{-2pt}
\begin{tabular}{rrr|rr|r}
\toprule
\multicolumn{3}{c}{Training Time} & \multicolumn{2}{c}{Validation} & \multicolumn{1}{c}{Test} \\
epochs & GPU-hours & days & loss & accuracy & pass rate \\
\midrule
27     & 91.5      & 2    & 8.65 & 0.497 & \textbf{0.267} \\
247    & 907.1     & 20   & \textbf{8.25} & \textbf{0.567} & 0.261 \\
\bottomrule
\end{tabular}
\addtolength{\tabcolsep}{2pt}
\caption{Statistics for two G2T-Anon checkpoints.
Training times do not account
for the four instances we restarted training from an earlier checkpoint.
Validation accuracy is the proportion of instances where the generated
base tactic plus arguments (using greedy decoding) 
match the ground truth from the held-out validation set.
The pass rate uses the G2T-Anon-Update configuration.}
\label{tab:training-time}
\end{table}

\begin{figure}[!htbp]
    \centering
    \includegraphics[width=1.0\columnwidth]{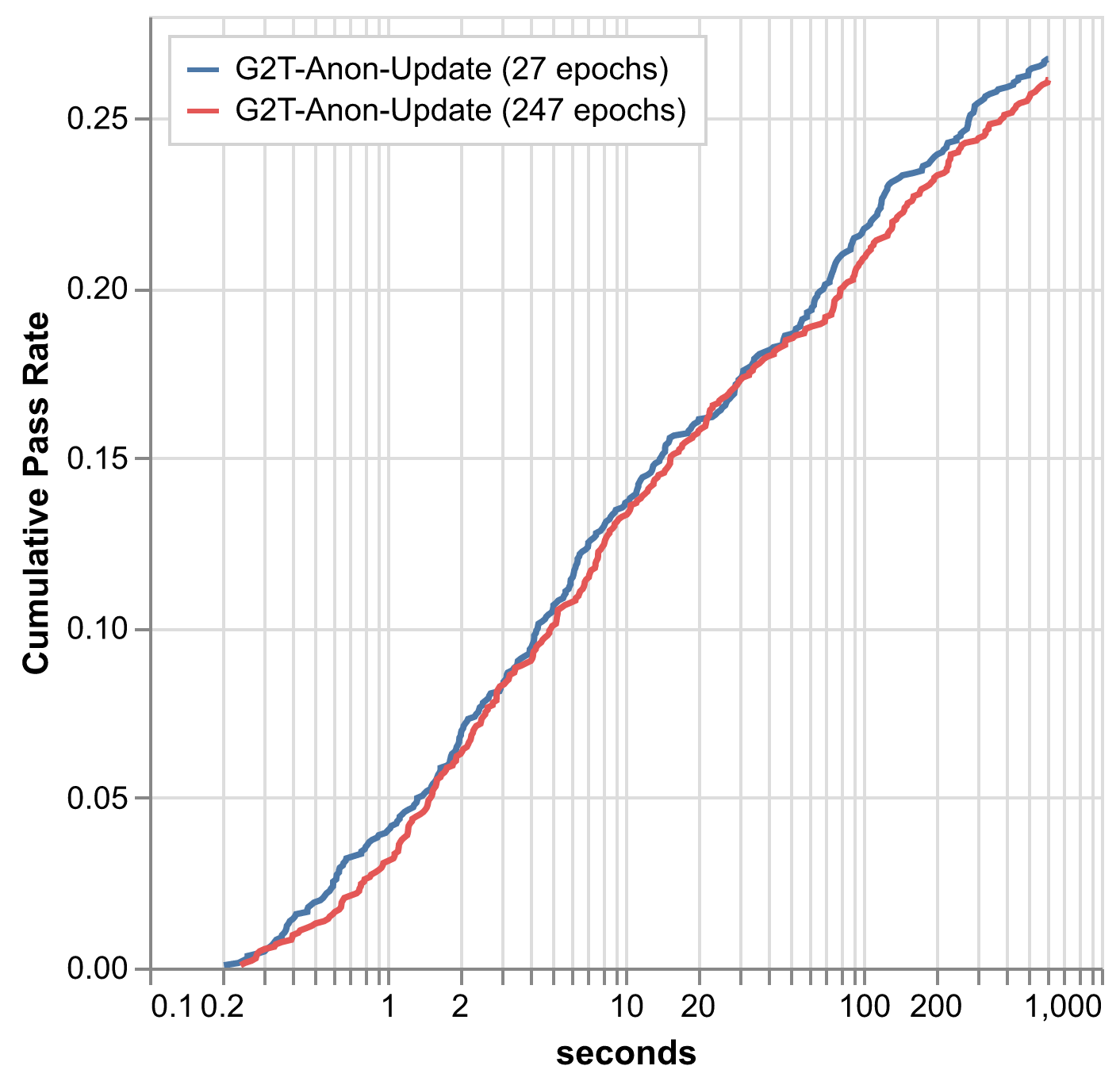}
    \caption{Pass rates for different checkpoints of G2T-Anon.}
    \label{fig:gt2-training-time}
\end{figure}

\section{Ablation of G2T Solvers and Symmetry Breaking with Names}\label{sec:g2t-ablation}

\begin{figure}[!htbp]
    \centering
    \includegraphics[width=1.0\columnwidth]{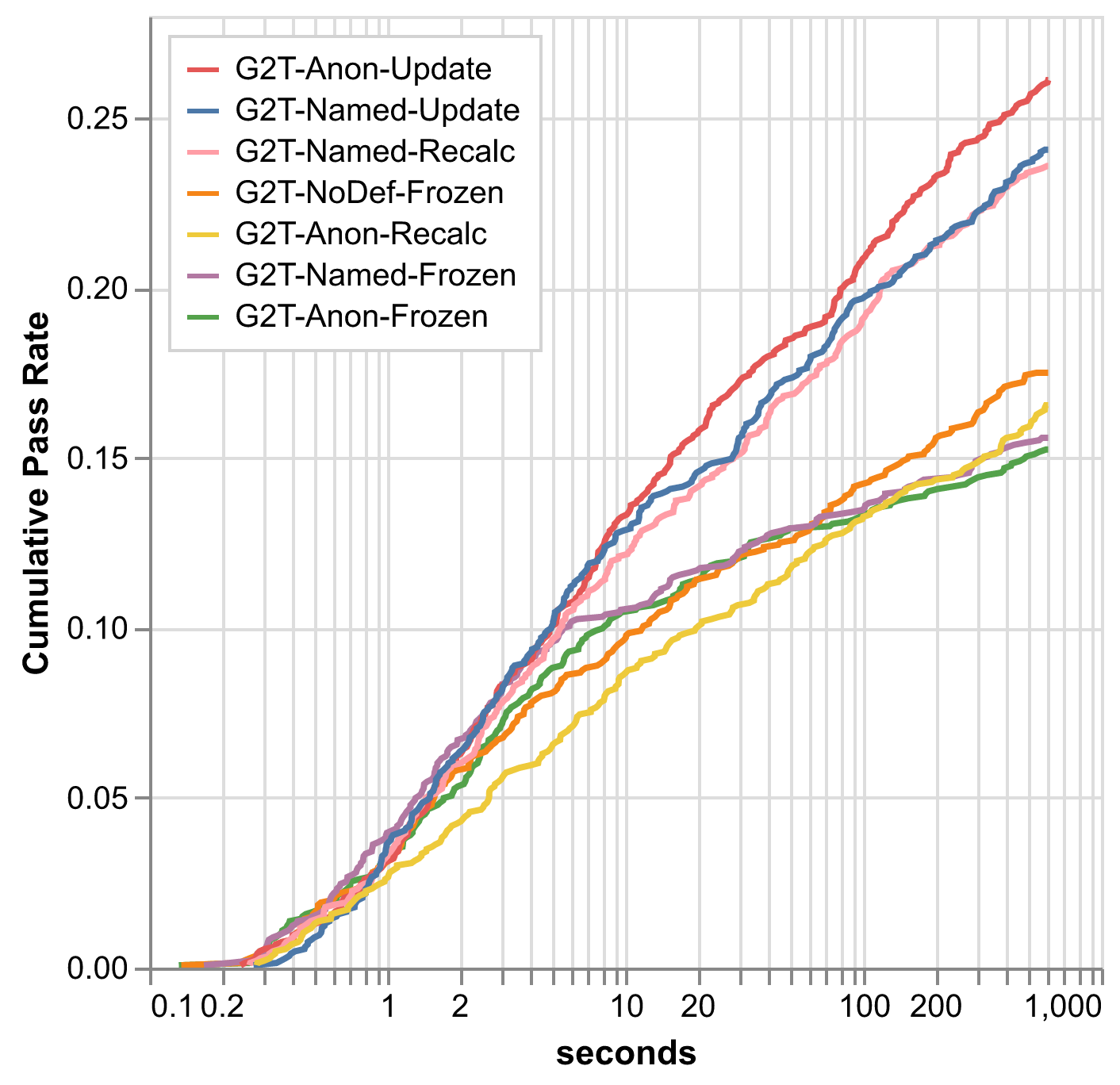}
    \caption{Pass rates for the different G2T solver variations.}
    \label{fig:gt2-test-packages-time}
\end{figure}

Figure~\ref{fig:gt2-test-packages-time} shows all seven G2T variations
mentioned in Subsection~\ref{sec:gnn-def-task}.
In particular, we see that using the frozen version of a model
where the definition task is not used to compute new definition embeddings
hurts model performance.
This suggests that our definition network is not just a helpful
auxiliary task during training,
but an essential component of our online system.

Also, note that the models with and without name embeddings have different
behaviors when we recalculate all definition embeddings
including those seen during training.
G2T-Named-Recalc shows the same performance as G2T-Named-Update.
Recalculating all embeddings would simplify the solver interface
since one does not have to keep
track of which definitions were part of training.

However, this is not true of G2T-Anon-Recalc.
One possible reason for this discrepancy is that
names break symmetries when calculating definition embeddings from scratch.
Isomorphic definitions have isomorphic graphs and hence receive the same embedding by our definition model (if name embeddings are not also used).
One example of this is the embeddings for the Boolean values \verb|true| and \verb|false|.
Further, this symmetry can propagate.
If \verb|true| and \verb|false| are given the same embedding,
then so are the Boolean operations \verb|andb| and \verb|orb|
since the graph of the former is isomorphic to the latter
with \verb|true| and \verb|false| swapped.

\section{Additional Combinations with the Online $k$-NN}
In our main results, we remark on how the online $k$-NN solver and
the G2T-Anon-Update solver are particularly complimentary.
We demonstrated this by plotting the combined time-scaled
aggregate in Figure~\ref{fig:test-packages-time}.
This is additionally supported by the Venn diagram in Figure~\ref{fig:venndiagram} which showed the G2T-Anon-Update
model and the online $k$-NN to be most disjoint.

\begin{figure}[!htbp]
    \centering
    \includegraphics[width=1.0\columnwidth]{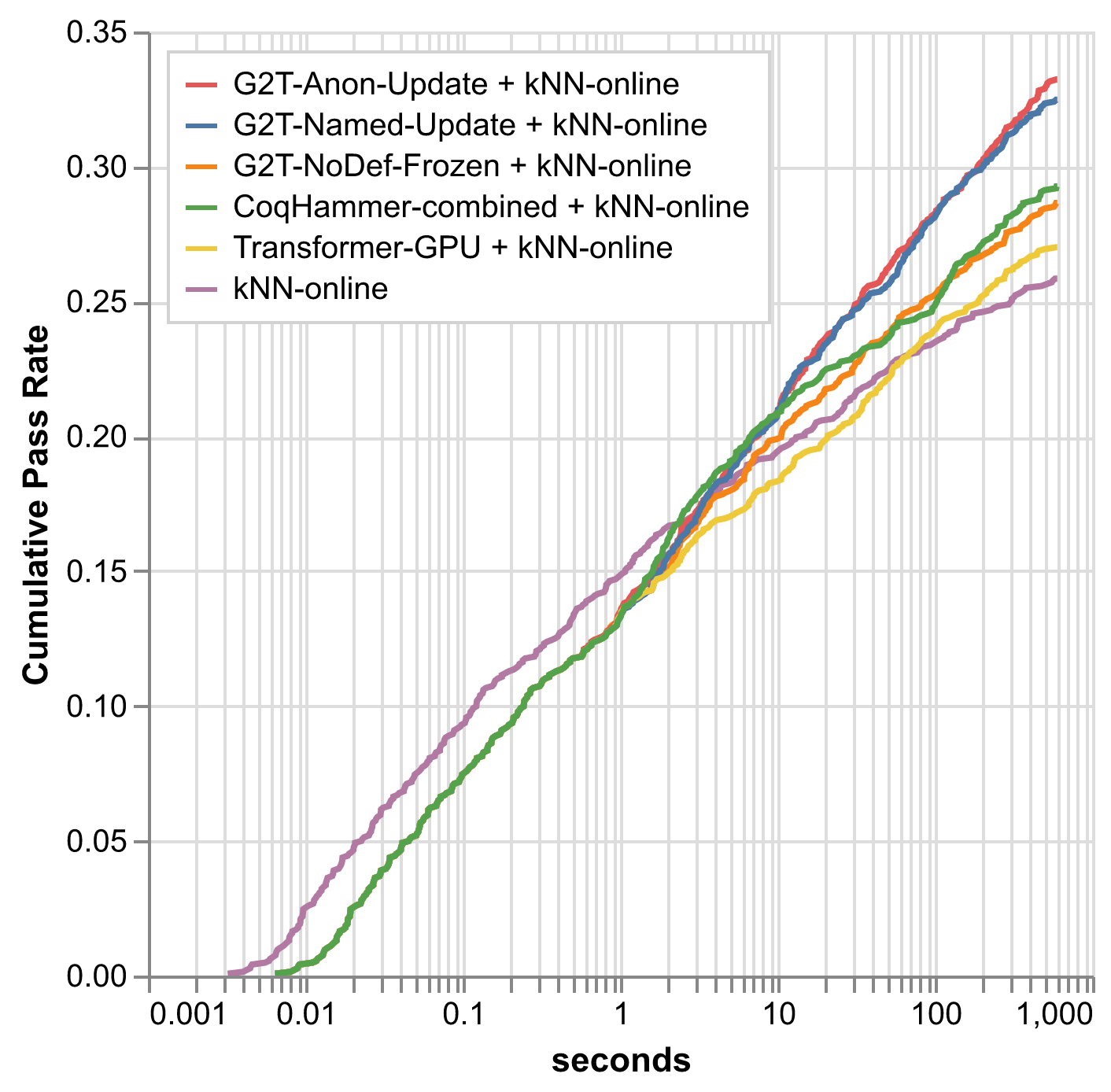}
    \caption{Pass rates of various solvers combined with the $k$-NN solver.  All combined solvers scale time so that time is evenly divided among the two component solvers.  The uncombined $k$-NN is provided for reference.}
    \label{fig:knn-combos-times}
\end{figure}

In Figure~\ref{fig:knn-combos-times} we further show that
of all the solvers, the online $k$-NN pairs best with
the G2T-Anon-Update and G2T-Named-Update models,
which both use our novel definition embeddings.
This again supports the complementary of the two online approaches,
as represented in Figure~\ref{fig:overview}.

\newpage

\section{Impact of the Online Setting}\label{sec:online-setting}
An important distinction of our paper is our use of
entire packages for testing, instead of splitting theorems into
test and train randomly as many other papers do.

To investigate this, we measure, for each of our test theorems,
the number of new definition dependencies
not seen during training
required for that theorem (both in its theorem statement and proof).
This includes transitive dependencies.

\begin{figure*}[!htbp]
    \centering
    \includegraphics[width=0.8\textwidth]{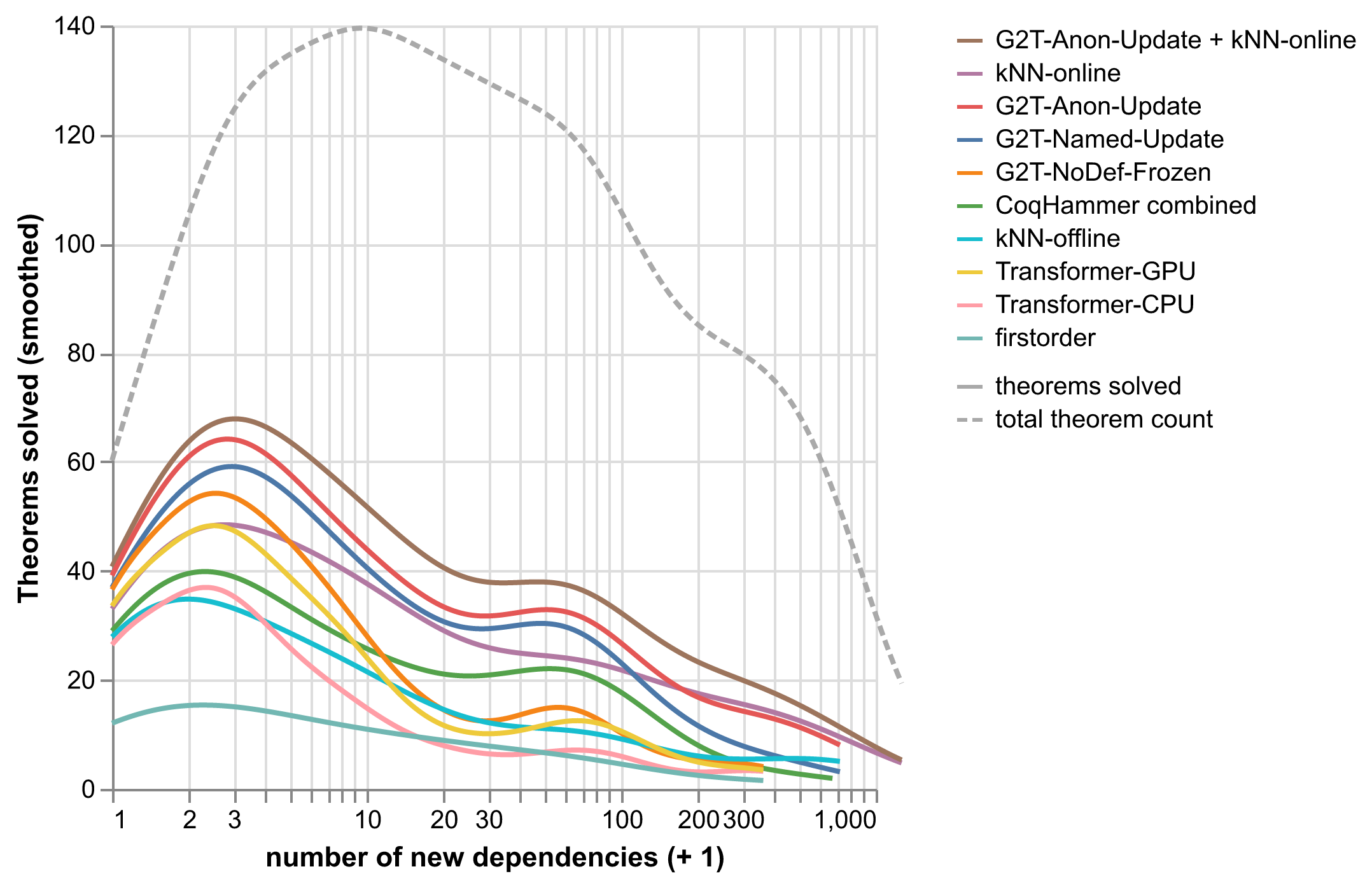}
    \caption{Theorems solved as a function of new dependencies.  (Smoothed to account for small numbers in each category.)  The dotted line shows the total number of test tactics (also smoothed) as some numbers of dependencies are more common.}
    \label{fig:dependencies-vs-passrate-plot}
\end{figure*}

\begin{figure*}[!htbp]
    \centering
    \includegraphics[width=1.0\textwidth]{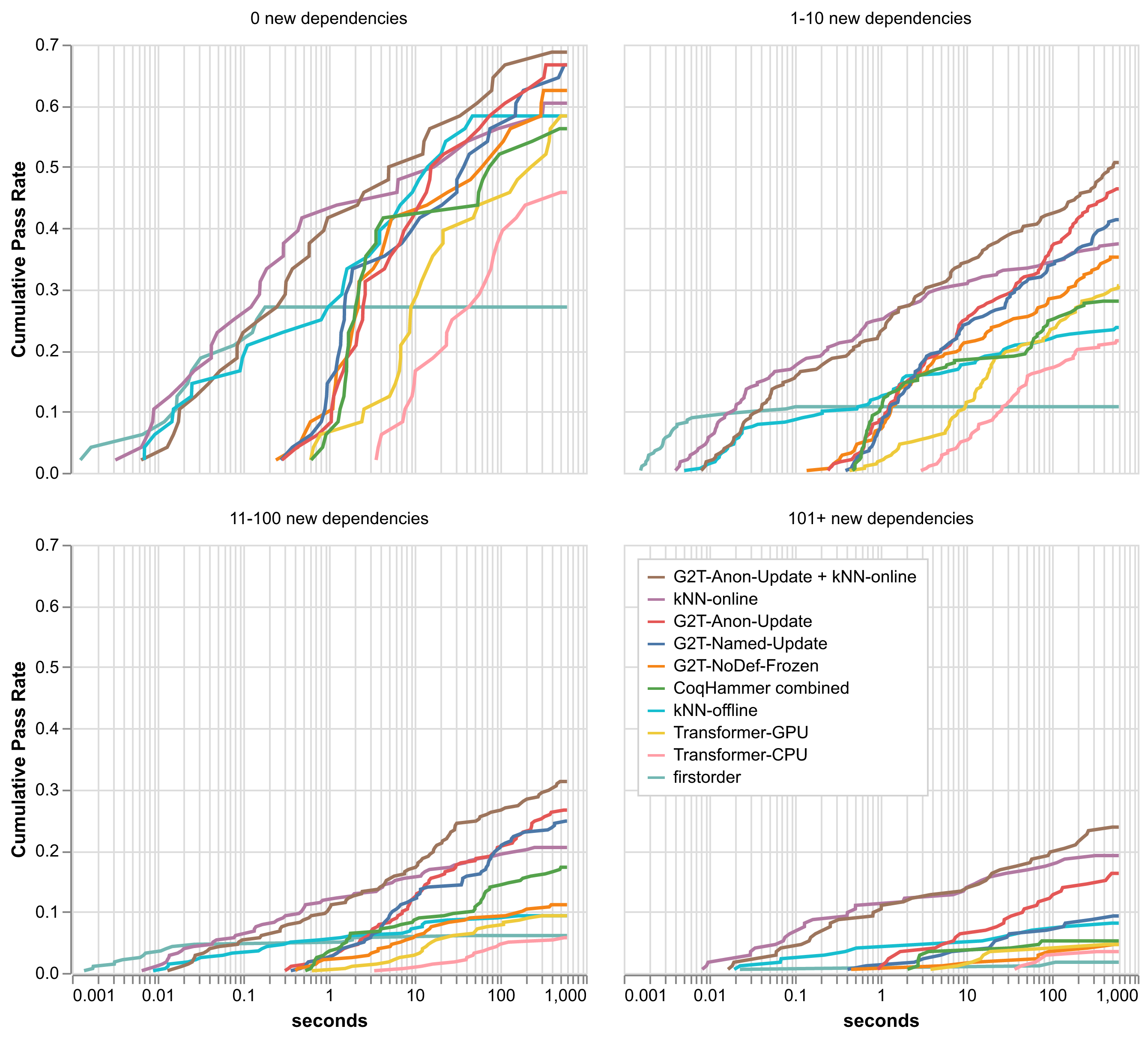}
    \caption{Cumulative pass rates as a function of time, split into four dependency categories.}
    \label{fig:dependencies-vs-passrate-grid}
\end{figure*}

\begin{figure*}[!htbp]
    \centering
    \includegraphics[width=1.0\textwidth]{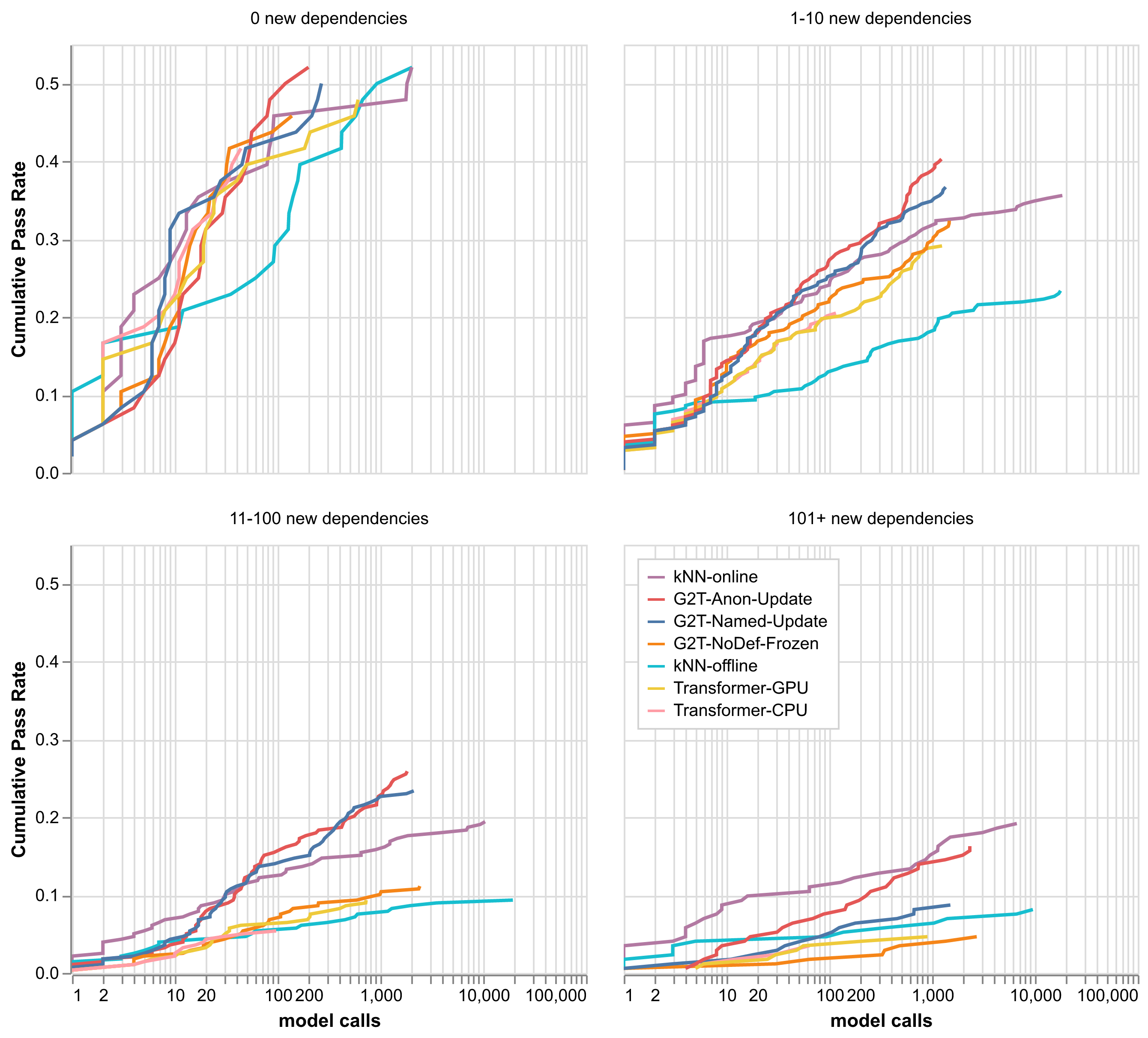}
    \caption{Cumulative pass rates as a function of model calls, split into four dependency categories.}
    \label{fig:dependencies-vs-passrate-grid-model-calls}
\end{figure*}

Figure~\ref{fig:dependencies-vs-passrate-plot} shows the number of solved theorems
as a function of new dependencies.
Figure~\ref{fig:dependencies-vs-passrate-grid} shows the per-time pass rate,
conditioned on the number of dependencies.
Figure~\ref{fig:dependencies-vs-passrate-grid-model-calls} is the same,
but conditioned on the number of model calls.%
\footnote{
Note that due to discrepancies in how we collect the dependency data,
not all test theorems are included in the figures.
}  In general, having more new dependencies is more challenging for all the solvers.

For 0 new dependencies, we see less difference between all the solvers,
and our best solvers solve over 60\% of theorems.
But when we move to a higher number of new dependencies,
the offline models,
including the transformer models, the offline $k$-NN, and G2T-NoDef-Update,
underperform the online models,
the online $k$-NN, G2T-Anon-Update, and G2T-Named-Update.
This is yet more justification that the main benefit of our definition task and the online $k$-NN is their usefulness in online settings.

Nonetheless, we do notice a small improvement in the G2T-Anon-Update
solver over the G2T-NoDef-Frozen solver in the 0 new dependency setting.
This suggests that the definition
training task may show some usefulness on its own as an auxiliary task
during training even when not used at inference time.

The online $k$-NN solver starts to outperform the G2T models when
dependencies get into the hundreds.
It is unclear why this happens.
In particular, there are many correlations
between the number of new dependencies and the package.
Larger Coq developments have a larger number of new dependencies,
but they may support some solvers over others for other orthogonal reasons.

\section{Discovery of Inconsistent Axioms}\label{sec:axioms}
After benchmarking our solvers we discovered that
three of our test packages contained inconsistent axioms.
These axioms are used for development or special logical manipulation
but are not intended to be used blindly in theorems,
as they can be used to prove false results.

\begin{table*}[!htbp]
\centering
\begin{tabular}{lllrr}
\toprule
package & axiom & solver & count & percent \\
\midrule
haskell & \verb|undefined| & G2T-Anon-Update & 3 & 1.7\% \\
haskell & \verb|undefined| & G2T-Named-Update & 8 & 4.6\% \\
haskell & \verb|undefined| & G2T-NoDef-Frozen & 13 & 7.5\% \\
hott & \verb|No_Empty_admitted| & G2T-Anon-Update & 9 & 1.8\% \\
hott & \verb|No_Empty_admitted| & $k$NN-online & 2 & 0.4\% \\
tlc & \verb|skip_axiom| & G2T-Anon-Update & 192 & 38.4\% \\
tlc & \verb|skip_axiom| & G2T-Named-Update & 5 & 1.0\% \\
tlc & \verb|skip_axiom| & G2T-NoDef-Frozen & 25 & 5.0\% \\
\bottomrule
\end{tabular}
\caption{The number of times the solvers use inconsistent axioms to prove a test theorem in the 5 minute benchmark.}
\label{tab:axiom-results}
\end{table*}

\begin{figure*}[!htbp]
    \centering
    \includegraphics[width=1.0\textwidth]{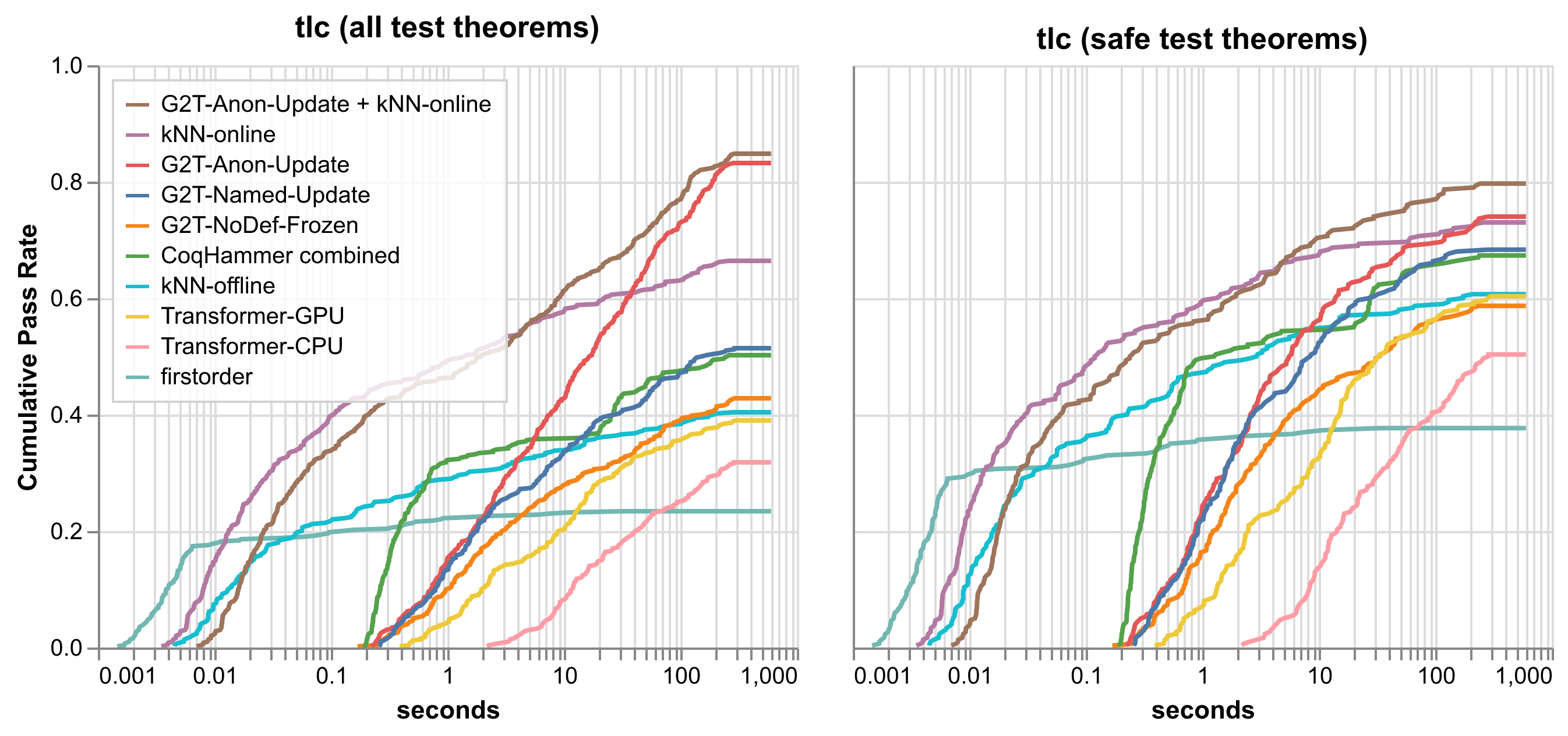}
    \caption{Pass rates on the \emph{tlc} test theorems.  The left plot shows the results on 500 theorems theorems.  The right plot restricts to test theorems not solved using an inconsistent axiom by any of the benchmarked solvers.}
    \label{fig:tlc-axiom-discussion-plot}
\end{figure*}

As shown in Table~\ref{tab:axiom-results},
some of our solvers were able to discover these axioms and use them.
The most significant case was \verb|skip_axiom| in the \emph{tlc} package.
G2T-Anon-Update managed to solve 192 of the 500 test theorems (38.4\%)
for the 5-minute benchmark using this axiom.
Figure~\ref{fig:tlc-axiom-discussion-plot} shows the results
with all test theorems
and when restricting to theorems for which no model used \verb|skip_axiom|.

It is impressive that our model was able to discover the
importance of this axiom,
but it also makes the evaluation of our results difficult.
For example, on \textit{tlc},
the name-less G2T model significantly
outperforms the one with names,
but mostly because it was able to exploit this axiom.

The axioms are as follows.  In \textit{tlc}:
\begin{minted}[fontsize=\small]{coq}
Axiom skip_axiom : False.
\end{minted}
In \textit{hott}:
\begin{minted}[fontsize=\small]{coq}
Inductive No_Empty_for_admitted := .
Axiom No_Empty_admitted
  : No_Empty_for_admitted.
\end{minted}
In \textit{haskell}:
\begin{minted}[fontsize=\small]{coq}
Definition undefined {a : Type} : a.
Admitted.
\end{minted}
In \textit{quickchick} (used for training):
\begin{minted}[fontsize=\small]{coq}
Axiom ignore_generator_proofs : False.
\end{minted}

\section{Examples of Theorems Solved}
Here we provide illustrative, cherry-picked examples
of how different solvers are able to exploit different types of information.

\paragraph{Using a New Definition in a Theorem}

In this example from \emph{odd-order},
the theorem \verb|nsKM| is in a Coq section and
depends on both the section variable \verb|M| and a new definition \verb|K|,
defined to be the product of the normal groups of \verb|M|.
\begin{minted}[fontsize=\small]{coq}
Variables M P0 : {group gT}.
Let K := M`_\F%G.

Let nsKM : K <| M.
Proof. exact: gFnormal. Qed.
\end{minted}

G2T-Anon-Update was the only model to find a proof, finding the same proof (on its first attempt)
as the original using the lemma \verb|gFnormal|.
\begin{minted}[fontsize=\small]{coq}
Lemma gFnormal gT (G : {group gT})
  : F gT G <| G.
\end{minted}

One possibility is that G2T-Anon-Update is using the hierarchical nature of our definition task,
as one has to go back a few steps in the definition hierarchy to verify that \verb|K|
fits the form \verb|F gT G| used in \verb|gFnormal|.

\paragraph{Using a New Lemma in a Proof}

In this next example from \emph{bits}, the lemma \verb|toZp_inj| uses
the immediately preceding new lemma \verb|toZpK|,
which was not seen during training.

\begin{minted}[fontsize=\small]{coq}
Lemma toZpK n : cancel (@toZp n) (@fromZp n).
Proof. case E: (n == 0).

Lemma toZp_inj n : injective (@toZp n).
Proof. apply (can_inj (@toZpK _)). Qed.
\end{minted}

G2T-Anon-Update and G2T-Named-Update can both solve this theorem with
a similar proof: \verb|apply : can_inj. exact : toZpK.|,
something neither the offline models or nor the online k-NN was capable of.

However, the the online k-NN can solve a similar example \verb|tiV| from \emph{bits},
because \verb|tiV| is the third in a list of facts, all using the same proof.

\begin{minted}[fontsize=\small]{coq}
Let cycW : cyclic W.
Proof. by case: ctiW. Qed.

Let oddW : odd #|W|.
Proof. by case: ctiW. Qed.

Let tiV : normedTI V G W.
Proof. by case: ctiW. Qed.
\end{minted}

The online k-NN can retrive that tactic \verb|by case: ctiW.| since it is in its database of recent proofs.
In this case, neither G2T-Anon-Update nor G2T-Named-Update could solve this, nor any of the offline solvers.
(However, CoqHammer's \verb|best| tactic was successful.)

\paragraph{Using New Tactics}

The online k-NN can also find and use new tactics, something none of the other solvers can do.
The package \emph{hott} is very different from the training data, 
often using custom tactics, like the tactic \verb|make_equiv| used in the proof of \verb|equiv_sigma_assoc'|.

\begin{minted}[fontsize=\small]{coq}
Definition equiv_sigma_assoc'
  `(P : A -> Type)
  (Q : forall a : A, P a -> Type)
  : {a : A & {p : P a & Q a p}}
    <~> {ap : sig P & Q ap.1 ap.2}.
Proof.
  make_equiv.
Defined.
\end{minted}

It was easy for the online k-NN to find this proof as most theorems in this file used the same tactic, while none of the other solvers could solve this.

\paragraph{Predicting Tactics with Terms}

While the transformer model rarely solved a proof that the wasn't also solvable by the G2T models,
there is an interesting example from \emph{hott}. The proof of \verb|meet_lb_3_assoc_m| relies on
constructing the term \verb|y ⊓ z|, something that,
of all the machine-learned solvers in this paper,
only the transformer is capable of.
\begin{minted}[fontsize=\small]{coq}
Lemma meet_lb_3_assoc_m x y z
  : x ⊓ (y ⊓ z) ≤ y.
Proof.
transitivity (y ⊓ z).
- apply meet_lb_r.
- apply meet_lb_l.
Qed.
\end{minted}
(This theorem was also proved by the firstorder tactic.)

\paragraph{Using Name Information}

Last, in \emph{tlc}, consider the theorem \verb|LibListZ.length_map|.

\begin{minted}[fontsize=\small]{coq}
Lemma length_map :
  forall A B (l:list A) (f:A->B),
  length (map f l) = length l.
Proof using.
  intros. unfold length. rewrite~ length_map.
Qed.
\end{minted}
Note, the proof of this theorem calls \verb|LibList.length_map|:
\begin{minted}[fontsize=\small]{coq}
Lemma length_map : forall f l,
  length (map f l) = length l.
\end{minted}
While it looks like these two theorems says the exact same thing,
the first theorem uses \verb|LibListZ.length| which gives an integer length
while the second uses \verb|LibList.length| which gives a natural number length.
Because G2T-Named-Update can see the name of the definitions, it had an advantage over
G2T-Anon-Update in this case and was able to solve \verb|LibListZ.length_map| with a similar proof,
while G2T-Anon-Update did not.

%Conversely, consider the ability of the G2T-Anon-Update model to exploit
%\verb|skip_axiom| from \emph{tlc} as seen in Section~\ref{sec:axioms},
%where as G2T-Named-Update did not
%\begin{minted}[fontsize=\small]{coq}
%    Axiom skip_axiom : False.
%\end{minted}
%This suggests G2T-Anon-Update can more effectively use
%the the axiom than the G2T-Named-Update model.
    
\section{Patterns in the Definition Embeddings}
\begin{figure*}[!htbp]
    \centering
    \includegraphics[width=1.0\textwidth]{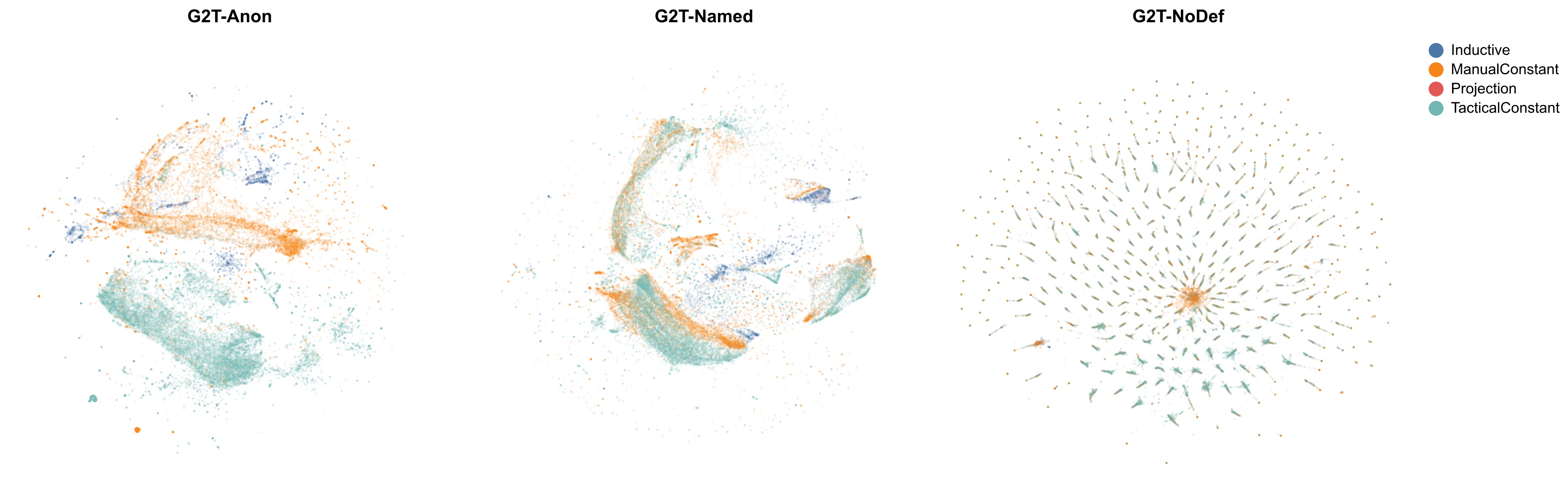}
    \caption{UMAP representations of learned definition embeddings
    colored by definition type.
    The G2T-Anon embeddings are clustered between
    tactical constants (mostly theorems with tactical proofs),
    manual constants (definitions defined via terms),
    and inductively defined definitions.
    (Projections in record types are relatively rare.)
    Manual and tactical constants are less separated in G2T-Named,
    but inductive definitions still do not mix.
    Manual constants make up a significant portion
    of the center cluster of the G2T-NoDef embeddings.}
    \label{fig:umap-def-type}
\end{figure*}

\begin{figure*}[!htbp]
    \centering
    \includegraphics[width=1.0\textwidth]{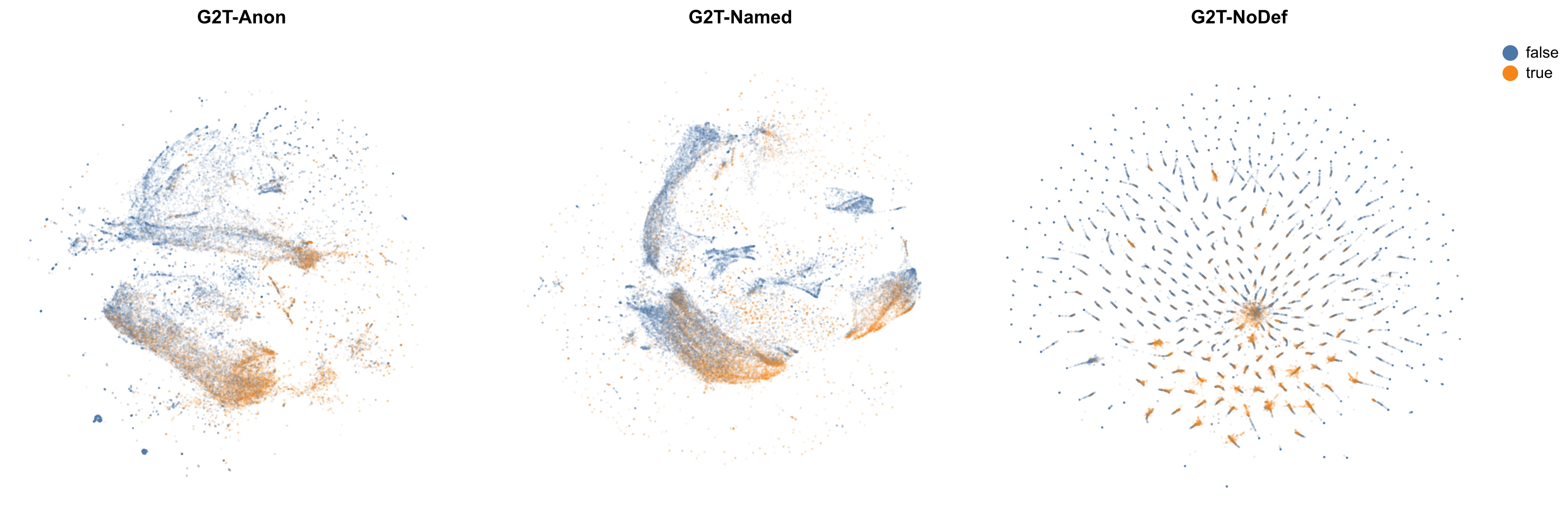}
    \caption{UMAP representations of learned definition embeddings
    colored by whether the definition was used as an argument of a tactic.
    All the model definition embeddings take this dimension into account.
    This makes sense since argument selection is an important part of tactic prediction.}
    \label{fig:umap-used-in-tactic}
\end{figure*}

\begin{figure*}[!htbp]
    \centering
    \includegraphics[width=1.0\textwidth]{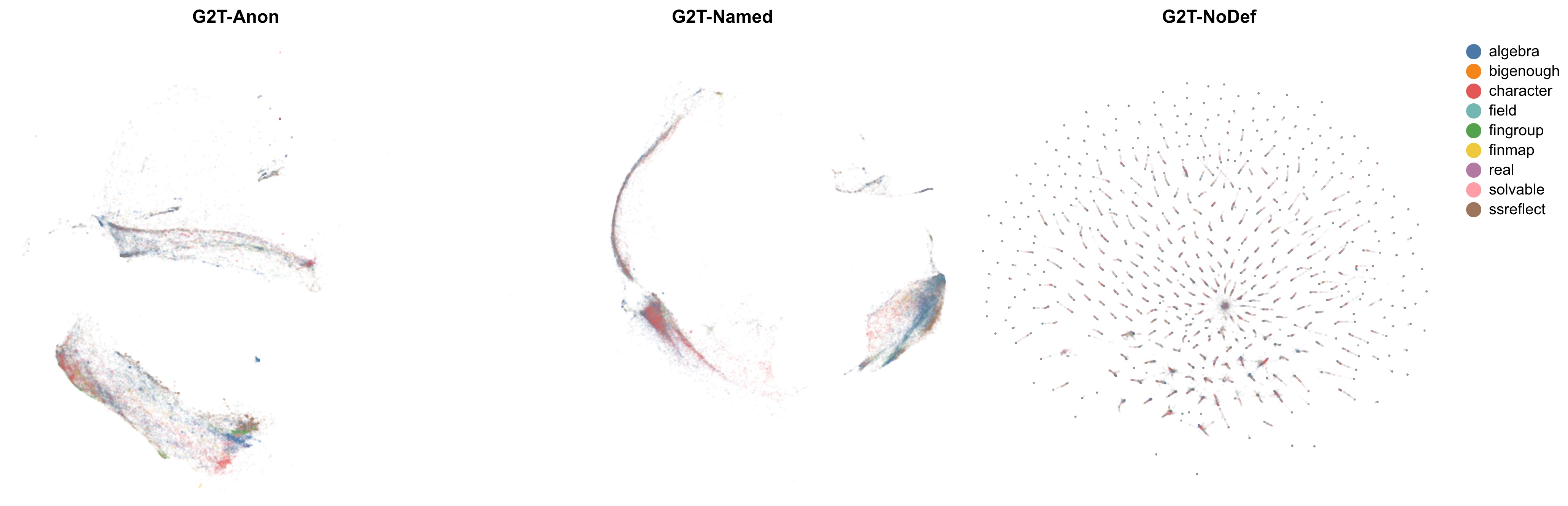}
    \caption{UMAP representations of learned definition embeddings
    restricted to definitions whose qualified names start with \texttt{mathcomp}
    colored by the second part of the name after \texttt{mathcomp},
    \emph{e.g.}~\texttt{mathcomp.\textbf{algebra}.matrix.Matrix}.
    The G2T-Named model clusters more by these categories,
    whereas in the G2T-Anon they show as smoothed-out bands.}
    \label{fig:umap-mathcomp-categories}
\end{figure*}

Our three models---G2T-Anon, G2T-Named, and G2T-NoDef---%
take different approaches in training the learned definition embeddings.
In Figures~\ref{fig:umap-def-type}, \ref{fig:umap-used-in-tactic}, and \ref{fig:umap-mathcomp-categories} we show UMAP representations of the learned definition embeddings.

Overall we see that G2T-Anon and G2T-Named have a more intricate structure.
Since G2T-Anon does not have access to names,
the embeddings are more focused on structure,
including the top-level
structure of what type of definition it is.

It is not clear what the major clusters of the G2T-Named model are,
although we can determine it is not just related to
easy-to-identify features of the name or the definition,
but instead seems to be a combination of multiple factors.

As for the G2T-NoDef, since there is no definition task,
the only signal used to train the embedding is how the definition is used, either as a node in a proof state or as an argument to a tactic.
The center seems to be almost exclusively made up of non-theorem definitions.
Further, we have noticed at least some of the radial clusters
are associated with particular tactics.
For example, we found a cluster almost exclusively
made up of definitions used as an argument to \verb|rewrite _|.

\end{document}